\documentclass{article} % For LaTeX2e
\usepackage{iclr2021_conference,times}

\usepackage{graphicx}
\usepackage{multirow}
\usepackage{booktabs}
\usepackage{bm}
\usepackage{balance}
\usepackage{lipsum}
\usepackage{enumitem}
\usepackage{amssymb}
\usepackage{amsmath}
\usepackage{mathrsfs}
\usepackage{xcolor}
\usepackage{xr}
\usepackage{textcomp}
\usepackage{wrapfig}
\usepackage{caption}

\newcommand{\Sref}[1]{\S\ref{#1}}

\newcommand{\Fref}[1]{Figure~\ref{#1}}

\newcommand{\Tref}[1]{Table~\ref{#1}}
\newcommand{\Aref}[1]{Appendix~\ref{#1}}

\usepackage{amsmath,amsfonts,bm}

\def\eqref#1{equation~\ref{#1}}
\def\1{\bm{1}}

\DeclareMathAlphabet{\mathsfit}{\encodingdefault}{\sfdefault}{m}{sl}
\SetMathAlphabet{\mathsfit}{bold}{\encodingdefault}{\sfdefault}{bx}{n}

\usepackage{hyperref}
\usepackage{url}

\newcommand{\method}{\textsc{DialoGraph}}
\newcommand{\structureencoder}{structure encoder}

\newcommand{\m}[1]{\mathcal{#1}}
\newcommand{\bmm}[1]{\bm{\mathcal{#1}}}
\newcommand{\real}[1]{\mathbb{R}^{#1}}

\title{\method{}: Incorporating Interpretable\\ Strategy-Graph Networks into\\ Negotiation Dialogues}

\author{Rishabh Joshi, Vidhisha Balachandran, Shikhar Vashishth, Alan W Black, Yulia Tsvetkov  \\
Language Technologies Institute \\
Carnegie Mellon University\\
\texttt{\{rjoshi2, vbalacha, svashish, awb, ytsvetko\}@cs.cmu.edu} %\\
}

\iclrfinalcopy % Uncomment for camera-ready version, but NOT for submission.
\begin{document}

\maketitle

\begin{abstract}
%Negotiations require complex skills of persuasive communication and pragmatic planning of effective negotiation strategies. While modern dialogue agents excel at generating fluent sentences, they still lack pragmatic grounding, and cannot reason strategically. We present \method{}, a graph-based negotiation system that models the sequence of strategies and their dependencies in negotiation dialogue using graph neural networks. \method{} explicitly incorporates dependencies between strategies to enable improved and interpretable prediction of a next optimal strategy, given the dialogue context. Our graph-based method outperforms prior state-of-the-art FST- and RNN-based  negotiation models on strategy and dialogue act prediction as well as the quality of downstream dialogue response generation. We further qualitatively demonstrate the benefit of our intermediate strategy-sequence graphs in interpreting the dialogue model decisions and understanding associations between negotiation strategies.\footnote{We will release the implementation, data and the negotiation dialogue demo upon acceptance of this work.}
  To successfully negotiate a deal, it is not enough to communicate fluently: pragmatic planning of persuasive negotiation strategies is essential. 
  While modern dialogue agents excel at generating fluent sentences, they still lack pragmatic grounding and cannot reason strategically. 
  We present \method{}, a negotiation system that incorporates pragmatic strategies in a negotiation dialogue using graph neural networks. 
  \method{} explicitly incorporates dependencies between sequences of strategies to enable improved and interpretable prediction of next optimal strategies, given the dialogue context. 
  Our graph-based method outperforms prior state-of-the-art negotiation models 
  %---which incorporate strategic context and dialogue act prediction using FSTs and RNNs---
  both in the accuracy of strategy/dialogue act prediction and in the quality of downstream dialogue response generation. 
  We qualitatively show further benefits of learned strategy-graphs in providing explicit associations between effective negotiation strategies over the course of the dialogue, leading to interpretable and strategic dialogues.\footnote{Code, data and a demo system is released at \url{https://github.com/rishabhjoshi/DialoGraph_ICLR21}}
  \end{abstract}

\section{Introduction}
\label{sec:introduction}

% \begin{figure}[ht]
%     %\centering
%     \begin{center}
%     \includegraphics[width=0.5\linewidth]{./images/motivation-crop.pdf}
%     \end{center}
%     \caption{An example of a negotiation dialogue between a buyer and a seller. Although both of the seller's options are equally plausible, the seller's utterance which incorporates better strategies in the current context (to mention family members and communicate informally) leads to a better deal (higher sale). 
%     }
%     \label{fig:intro_example}
% \end{figure}

Negotiation is ubiquitous in human interaction, from e-commerce to the multi-billion dollar sales of companies. 
Learning how to negotiate effectively involves deep pragmatic understanding and planning the dialogue strategically \citep{thompson2001mind,thompsonpaper,pruitt2013negotiation}.

\begin{wrapfigure}{R}{0.40\textwidth}
    \vspace{-4mm}
    \centering
    \includegraphics[width=\linewidth]{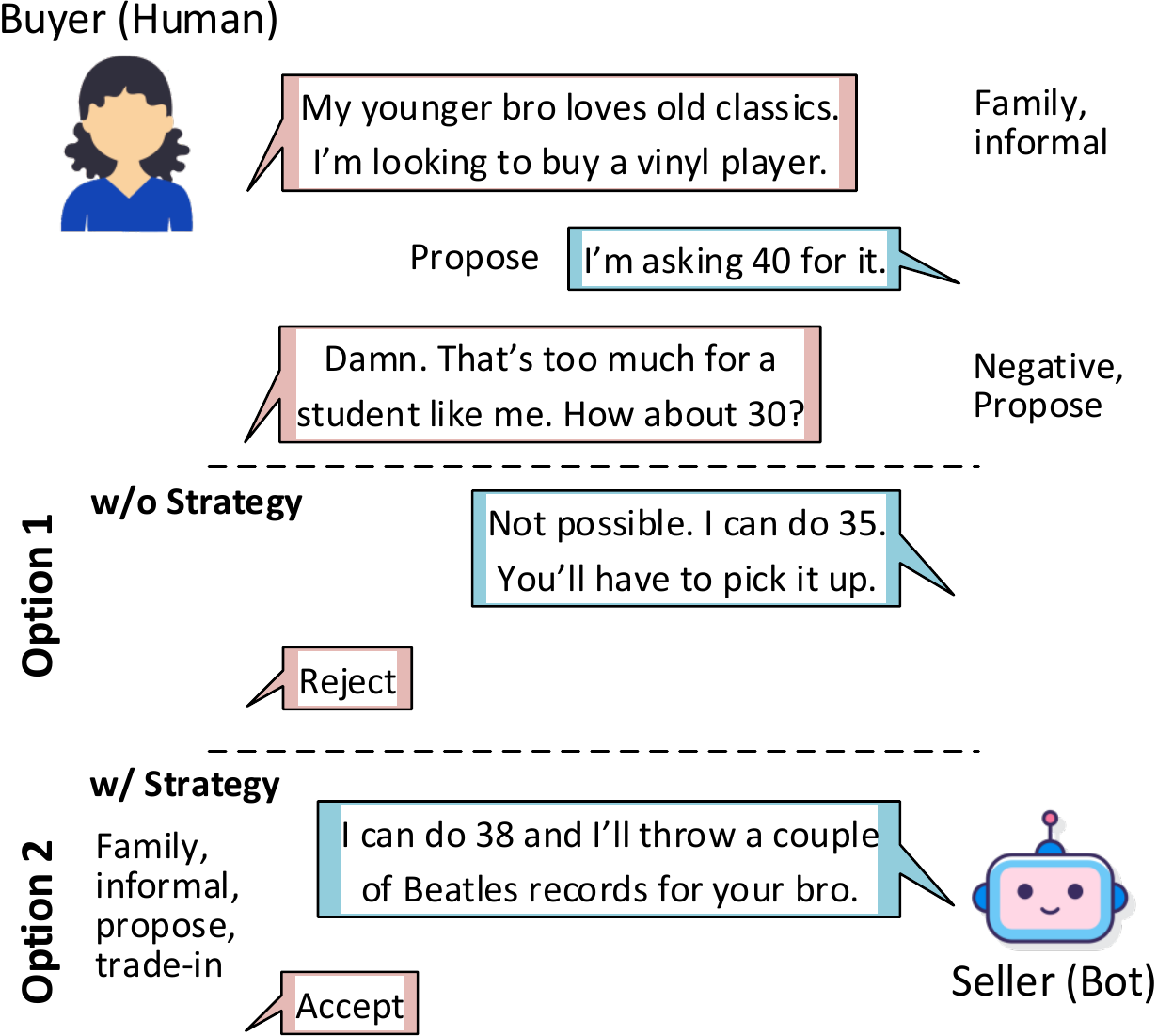}
    \caption{\small{Both options are equally plausible and fluent, but a response with effective pragmatic strategies %(family member mentions and informal communication) 
    leads to a better deal.\\}} \vspace{-4mm}
    \label{fig:intro_example}
\end{wrapfigure}
Modern dialogue systems for collaborative tasks such as restaurant or flight reservations have made considerable progress by modeling the dialogue history and structure explicitly using the semantic content, like slot-value pairs \citep{tartan2018,youngcollab}, or implicitly with encoder-decoder architectures \citep{sordoni-etal-2015-neural,li-etal-2016-persona}. 
In such tasks, users communicate explicit intentions, enabling systems to map the utterances into specific intent slots \citep{li_endtoend}. 
However, such mapping is less clear in complex non-collaborative tasks like \emph{negotiation} \citep{negotiation_og} and \emph{persuasion} \citep{persuasion4good}, where user intent and most effective strategies are hidden. 
Hence, along with the generated dialogue, the strategic choice of framing and the sequence of chosen strategies play a vital role, as depicted in \Fref{fig:intro_example}. %\textcolor{blue}{Here, although both of the seller's options are equally plausible, the seller's utterance which incorporates better strategies in the current context (to mention family members and communicate informally) leads to a better deal (higher sale).} % for building effective negotiation dialogue systems. % \citep{thompson2001mind,thompsonpaper}. 
Indeed, prior work on negotiation dialogues has primarily focused on optimizing dialogue strategies---from high-level task-specific strategies \citep{dealornodeal}, to more specific task execution planning \citep{negotiation_og}, to fine-grained planning of linguistic outputs given strategic choices \citep{yiheng_sigdial}.
These studies have confirmed that it is crucial to control for pragmatics of the dialogue to build effective negotiation systems. 

To model the explicit dialogue structure, prior work incorporated Hidden Markov Models (HMMs) \citep{zhai-williams-2014-discovering,ritter-etal-2010-unsupervised}, Finite State Transducers (FSTs) \citep{yiheng_iclr} and RNNs \citep{negotiation_og,shi-etal-2019-unsupervised}. 
While RNN-based models lack interpretability, HMM- and FST-based approaches may lack expressivity.  %and thus do not adequately model complex negotiation settings. 
%More recently, there has been an increasing interest in utilizing 
In this paper, we hypothesize that Graph Neural Networks (GNNs) \citep{gnn_survey} 
can combine the benefits of interpretability and expressivity 
because of their effectiveness in encoding graph-structured data through message propagation. While being sufficiently expressive to model graph structures, GNNs also provide a natural means for interpretation via intermediate states \citep{gnn_interpret1,gnn_interpret2}.

We propose \method{}, an end-to-end negotiation dialogue system that leverages Graph Attention Networks (GAT) \citep{gat} to model complex negotiation strategies while providing interpretability for the model via intermediate structures. 
%\method{} incorporates the recently proposed hierarchical graph pooling based approaches \citep{asap} to learn the associations between- and the relative importance of negotiation strategies---including conceptual and linguistic strategies, as well as dialogue acts---in predicting the best sequence. 
\method{} incorporates the recently proposed hierarchical graph pooling based approaches \citep{asap} to learn the associations between negotiation strategies, including conceptual and linguistic strategies and dialogue acts, and their relative importance in predicting the best sequence.
We focus on buyer--seller negotiations in which two individuals negotiate on the price of an item through a chat interface, and we model the seller's behavior on the CraigslistBargain dataset \citep{negotiation_og}.\footnote{We focus on the seller's side following \citet{yiheng_sigdial} who devised a set of strategies specific to maximizing the seller's success. Our proposed methodology, however, is general. } 
We demonstrate that \method{} outperforms previous state-of-art methods on strategy prediction and downstream dialogue responses.
This paper makes several contributions.  First, we introduce a novel approach to model negotiation strategies and their dependencies as graph structures, via GNNs. Second, we incorporate these learned graphs into an end-to-end negotiation dialogue system and demonstrate that it consistently improves future-strategy prediction and downstream dialogue generation, leading to better negotiation deals (sale prices). Finally, we demonstrate how to interpret intermediate structures and learned sequences of strategies, opening-up the black-box of end-to-end strategic dialogue systems.
%itemsep=1pt ,parsep=0pt,partopsep=0pt,leftmargin=1pt,topsep=1pt
%\begin{itemize}[itemsep=1pt ,parsep=0pt,partopsep=0pt,leftmargin=10pt,topsep=1pt]
%    \setlength\itemsep{1pt}
%    \setlength\parsep{0pt}
%    \setlength\partopsep{0pt}
%    \setlength\leftmargin{10pt}
%    \setlength\topsep{1pt}
%	\item We propose \method{} which uses Graph Neural Networks to model negotiation strategies and their dependencies as graph structures.
%	\item We incorporate \method{} into a negotiation dialogue system and demonstrate that it consistently improves future-strategy and downstream dialogue generation leading to better negotiation deals (sale price).
%	\item We qualitatively demonstrate the ability of \method{} to provide intermediate structures for interpretation.
% 	\item We propose a modularized framework that can be applied to other tasks to learn the dialogue structure better. \reminder{should i mention this? given that i have not done on other tasks. and not many other tasks involve multiple labels per utterance kind of structure}
%\end{itemize}

\section{\method{}} % Overview}
\begin{figure*}[t]
	%\centering
	\begin{center}
	\includegraphics[width=0.70\textwidth]{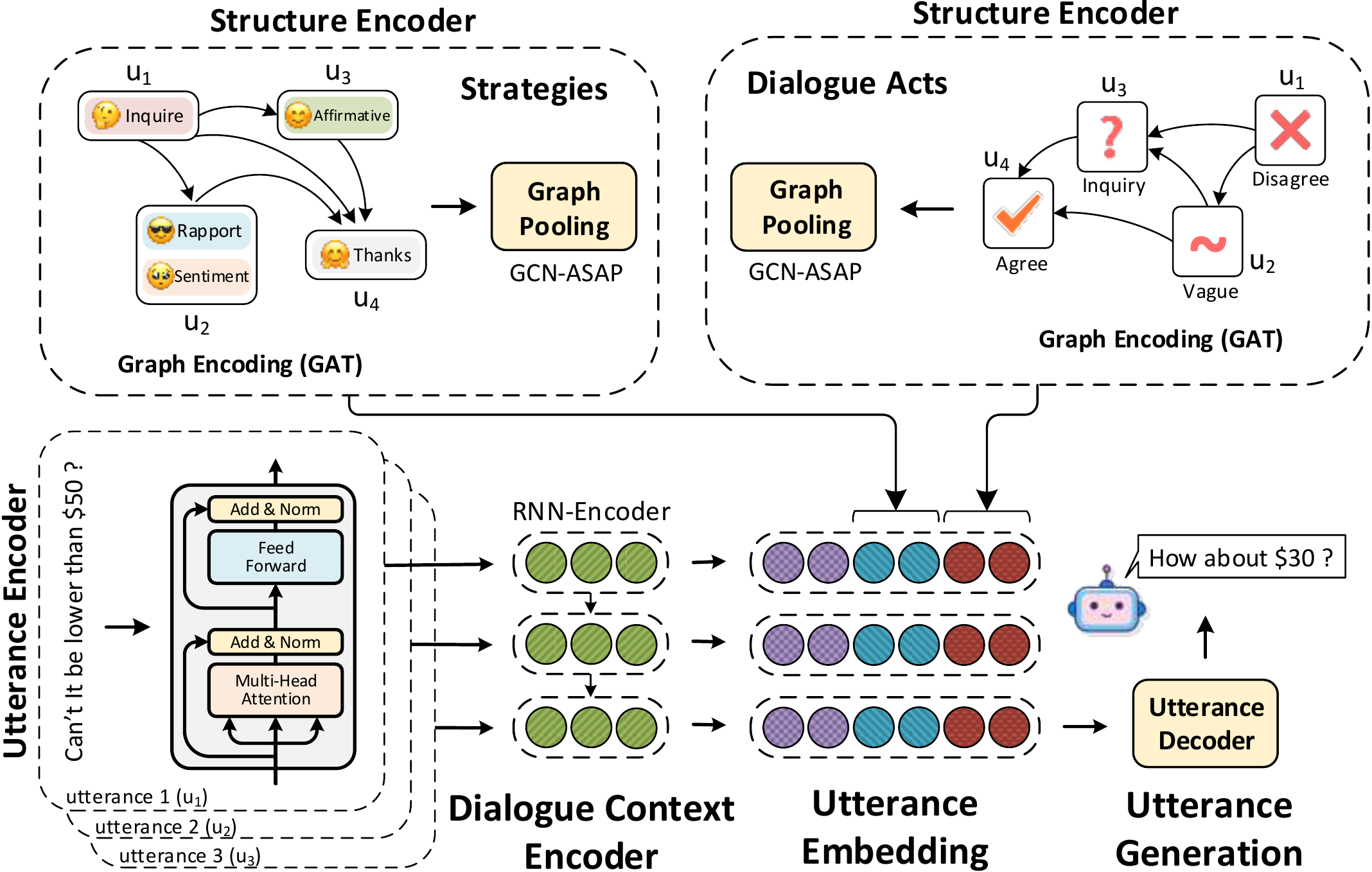}
	\end{center}
	%%\caption{\label{fig:workflow} \method{} Model Architecture. 
    \caption{\label{fig:workflow} Overview of \method{}. At time $t$, utterance $u_t$ is encoded using BERT and then passed to the Dialogue Context Encoder to generate the dialogue representation. This representation is enriched with the encodings of explicit strategy and dialogue act sequences using the \structureencoder{}s which is then used to condition the Utterance decoder. Please refer to \Sref{sec:method_details} for details.}
\end{figure*}

\label{sec:method_details}
We introduce \method{}, a  modular end-to-end dialogue system, that incorporates GATs with hierarchical pooling to learn pragmatic dialogue strategies jointly with the dialogue history. 
\method{} is based on a hierarchical encoder-decoder model and consists of three main components: (1) \emph{hierarchical dialogue encoder}, which learns a representation for each utterance and encodes its local context; (2) \emph{\structureencoder{}} for encoding sequences of negotiation strategies and dialogue acts; and (3) \emph{utterance decoder}, which finally generates the output utterance.
Formally, our dialogue input consists of a sequence of tuples, $\m{D} = [(u_1, da_1, ST_1), (u_2, da_2, ST_2), ..., (u_n, da_n, ST_n) ]$ where $u_i$ is the utterance, $da_i$ is the coarse dialogue act and $ST_i = \{st_{i,1}, st_{i,2}, \ldots, st_{i,k}\}$ is the set of $k$ fine-grained negotiation strategies for the utterance $u_i$.\footnote{For example, in an utterance \textit{Morning! My bro destroyed my old kit and I'm looking for a new pair for \$10}, the coarse dialogue act is \textit{Introduction}, and the finer grained negotiation strategies include \textit{Proposing price}, \textit{Being informal} and \textit{Talking about family for building rapport}.}  The dialogue context forms the input to (1) and the previous dialogue acts and negotiation strategies form the input to (2).
%
%We pass all the utterances through pre-trained BERT model \citep{bert} and use the pooled outputs as input to a GRU-based dialogue context encoder to obtain a representation of the conversation. 
%The two \structureencoder{}s then use the graph representations of the strategies and dialogue acts to give structural representations of the conversation which is used to predict the next set of strategies and dialogue acts (refer to Section \ref{sec:graph_details}).
%These representations are used to enrich the dialogue representation which is then used as input to the Utterance Decoder to generate the next utterance. 
%The final enriched dialogue representation is used to predict the conversation success outcome which represents the sale-to-price ratio. 
%\textcolor{blue}{In addition to that, we also generate the next utterance $u_{n+1}$ for the negotiation system which is conditioned on the predicted strategies and dialogue acts.} 
The overall architecture is shown in \Fref{fig:workflow}. 
%The focus of this work is to model and predict the set of strategies and dialogue acts, $ST_{n+1}$ and $da_{n+1}$ for the next utterance \textcolor{blue}{and hence we borrow the utterance decoder from \citet{yiheng_iclr}}.
%Following them, we model the seller's behaviour in negotiation and consequently, make the predictions and generations for the seller. 
In what follows, we describe \method{} in detail. %, the Hierarchical Dialogue Encoder, Sequence Graph Encoder, and the Utterance Decoder.

\subsection{hierarchical dialogue encoder}
%\noindent \textbf{Hierarchical Dialogue Encoder: }
\label{sec:model_encoder}
A dialogue context typically comprises of multiple dialogue utterances which are sequential in nature. We use hierarchical encoders for modeling such sequential dialogue contexts \citep{higru}. To encode the utterance $u_t$ at time $t$, we use the pooled representations from BERT \citep{bert} to obtain the corresponding utterance embedding $\bm{e}_t$. 
We then pass the utterance embeddings through a GRU to obtain the dialogue context encoding till time $t$, denoted by $\bm{h}_t^{U}$.

%\begin{figure*}[t]
%	\centering
%	\includegraphics[width=0.65\textwidth]{./images/strategy_flow-crop.pdf}
%	\caption{\label{fig:graph_creation} Visualization of a strategy sequence graph. The graph connects each strategy with all previously occurring strategies. Here we present only a few edges for brevity. For example, there would be two more additional edges from $u_4$ to the strategies of $u_5$. \reminder{subset of edges might be misleading}
%}
%\end{figure*}

\subsection{\structureencoder{}}
%\noindent \textbf{\structureencoder{}: }
\label{sec:graph_details}
Our \structureencoder{} is designed to model the graph representations of the strategies and dialogue acts using GATs and output their structural representations. These structural representations are used to predict the next set of strategies and dialogue acts and enrich the encoded dialogue representation. Below we describe the \structureencoder{} for negotiation strategies.

We model the sequence of negotiation strategies, $ST = [ ST_1, ST_2, \ldots, ST_t ]$ by creating a directed graph, where $ST_i$ is the set of k fine-grained negotiation strategies for the utterance $u_i$. Formally, we define a graph $\m{G}(\m{V}, \m{E}, X)$ with $| \m{E} |$ edges and $N = | \m{V} |$ nodes where each node $v_i \in \m{V}$ represents a particular negotiation strategy for an utterance and has a $d$-dimensional feature representation denoted by $\bm{z}_i$.
$\bm{Z} \in \real{N \times d}$ denotes the feature matrix of the nodes %node feature matrix
and $\bm{A} \in \real{N \times N}$ represents the adjacency matrix, where $N$ is the total number of nodes (strategies) that have occurred in the conversation till that point.
Therefore, each node represents a strategy-utterance pair. 

%Motivated by the success of direct connections in terms of residual connections \citep{resnet} and self-attention \citep{self_attention_parikh,transformer}, 
We define the set of edges as $\m{E} = \{(a,b)\} ; a, b \in \m{V}$ where $a$ and $b$ denote strategies at utterances $u_a$ and $u_b$, present at turns $t_a$ and $t_b$, such that $t_b > t_a$. % \reminder{Motivation to represent strategy as graph should not come from attention and residual. Some inconsistency here}.
In other words, we make a directed edge from a particular node (strategy in an utterance) to all the consecutive nodes. This ensures a direct connection from all the previous strategies to the more recent ones.\footnote{\Aref{sec:appendix_graph_creation} shows an example of the graph obtained from a sequence of strategies.} 
%We are motivated by the success of direct connections in terms of residual connections \citep{resnet} and self-attention \citep{attention} and hypothesize that having direct edges would enable our model learn the future tactics better. 
%
%\reminder{Vdeleted:In Section \ref{sec:next_strategy_prediction}, we show that such a direct connection indeed helps to improve the next-strategy-prediction performance. 
%We also show that connecting the just previous turn's tactics is not comparable with having all the history connected directly.} 
In the same way, we form the graph out of the sequence of dialogue acts. These direct edges and learned edge attention weights help us interpret the dependence and influence of strategies on each other.%flow of information.

% \noindent \textbf{Structure Encoder Model: }
To get the structural representations from the strategy graphs, we pass them through a hierarchical graph pooling based encoder, which consists of $l$ layers of GAT, each followed by the Adaptive Structure Aware Pooling (ASAP) layer \citep{asap}. 
As part of the ASAP layer, the model first runs GAT over the input graph representations to obtain structurally informed representations of the nodes. 
Then a cluster assignment step is performed which generates a cluster assignment matrix, $S$, which tells the model which nodes come in a similar structural context. 
After that, the clusters are ranked and then the graph is pooled by taking the top few clusters as new nodes and forming edges between them using the existing graph.
This way the size of the graph is reduced at every step which leads to a structurally informed graph representation.
We take advantage of the cluster formulation to obtain the associations between the negotiation strategies, as identified from the cluster assignment matrix, $S$.
These association scores can later be used to interpret which strategies are associated with each other and tend to co-occur in similar contexts.
%%The clusters are further ranked using the Local Extrema Convolution (LEConv) operator \citet{asap}, which gives the fitness score $\phi_i$ of each cluster, which essentially represents the ranking score that is used to pool the graph, by choosing important clusters.
%%These can be used to interpret the relative importance of different negotiation tactics to the future predictions. 
%%%%%This is in contrast to Transformers, which can only tell us what previous utterances were attended to. 
Moreover, we also use the node attention scores from GAT to interpret the influence of different strategies on the representation of a particular strategy, which essentially gives the dependence information between strategies. 
%We direct interested readers to \citet{asap} for more details regarding their ASAP Pooling layer. 

In this way, the structure representation is learned and accumulated in a manner that preserves the structural information \citep{diffpool,sagpool}.
After each pooling step, the graph representation is summarized using the concatenation of \textit{mean} and \textit{max} of the node representations. 
The summaries are then added and passed through fully connected layers to obtain the final structural representation of the strategies $\bm{h}_t^{ST}$. 
We employ a similar \structureencoder{} to encode the graph obtained from the sequence of dialogue acts, to obtain $\bm{h}_t^{da}$. 
%\textcolor{blue}{Here we describe the ASAP pooling layer formally. Initially each node $v_i$ in the graph is considered as a medoid of a cluster $c_h(v_i)$. We then obbtain the master query $m_i = \max_{v_j \in c_h(v_i)} (x_j^{'}))$ where $x_j^{'}$ is obtained after passing $x_j$ through a GAT to capture structural information in cluster $c_h(v_i)$. The master query $m_i$ attends to all constituent nodes $v_j \in c_h(v_i)$ using additive attention, $\alpha_{i,j} = \operatorname{softmax}(\mathbf{w}^T\sigma(Wm_i || x_j^'))$ where $\mathbf{w}^T$ and $W$ are learnable vector and matrix respectively. $\alpha_{i,j}$ signifies membership strength of node $v_j$ in cluster $c_h(v_i)$ and hence $S_{i,j} = \alpha_{i,j}$. Cluster representation is then $x^c_i = \sum_{j=1}^{|c_h(v_i)|} \alpha_{i,j}x_j$.
%The clusters are then scored using Local Extrema Convolution (LEConv) to compute $\hat{X} = \operatorname{LEConv}(X^c)$ and the $\operatorname{TOP}_k(.)$ function is used to rank the fitness scores obtained using LEConv to obtained the pooled graph given by $\hat{S} = S(:,\hat{i})$ and $X_p = \hat{X}^c(\hat{i},:)$ where $\hat{i} = \operatorname{TOP}_k(\hat{X}^c, k)$ is used for index slicing.
%The new adjacency matrix is then give by $A^p = \hat{S}^T\hat{A}^c\hat{S}$ where $\hat{A} = A^c + I$ and $A^c$ is the adjacency matrix of the graph. 
%$\phi = \sigma(x_i^cW_1 + \sum_{j \in N(i)} A^c_{i,j}(x^c_iW_2 - x^c_jW_3))$ where $N(i)$ is the neighborhood of node $i$ and $W_{1/2/3}$ are learnable parameters with $\sigma$ being an activation function. 
%}

\subsection{utterance decoder} 
\label{sec:decoder}
%\noindent \textbf{Utterance Decoder: }
The utterance decoder uses the dialogue context representation and structural representations of dialogue acts and negotiation strategies to produce the dialogue response (next utterance). We enrich the dialogue representation by concatenating the structural representations before passing it to a standard greedy GRU \citep{gru_paper} decoder. This architecture follows \citet{yiheng_iclr}, who introduced a dynamic negotiation system that incorporates negotiation strategies and dialogue acts via FSTs. We thus follow their utterance decoder architecture to enable direct baseline  comparison.   
For the $j^{th}$ word of utterance $u_{t+1}$, $w_{t+1}^{j}$, we condition on the previous word $w_{t+1}^{j-1}$ to calculate the probability distribution over the vocabulary as $\bm{p}_{t+1}^{w_j} = ~\  \mathrm{softmax}(\mathrm{GRU}(\bm{h}_t, \bm{w}^{j-1}_{t+1}))$ where $\bm{h}_t = ~ [\bm{h}_t^u ; \bm{h}_t^{ST} ; \bm{h}_t^{da}]$ and $[ ; ]$ represents the concatenation operator. For encoding the price, we replace all price information in the dataset with placeholders representing the percentage of the offer price. For example, we would replace $\$35$ with $<price-0.875>$ if the original selling price is $\$40$. The decoder generates these placeholders which are then replaced with the calculated price before generating the utterance.

%We condition on the previous word to calculate the probability distribution $\bm{p}_j^{t+1}$ over the vocabulary at time $j$ as $\bm{p}_j^{t+1} = ~\  \mathrm{softmax}(\mathrm{GRU}(\bm{h}_t, \bm{w}_{j-1}^{t+1}))$ where $\bm{h}_t = ~ [\bm{h}_t^d ; \bm{h}_t^g(st) ; \bm{h}_t^g(da)]$. %: 
% \begin{equation*}
% \label{eqn:nlg}
% %\begin{split}
% \bm{h}_t = ~ [\bm{h}_t^d ; \bm{h}_t^g(st) ; \bm{h}_t^g(da)] ~~~ ~~~~~\bm{p}_j^{t+1} = \  \mathrm{softmax}(\mathrm{GRU}(\bm{h}_t, \bm{w}_{j-1}^{t+1})).
% %\end{split}
% \end{equation*}
% \begin{equation*} % COMMENTED THIS
% \label{eqn:nlg}
% \begin{split}
% \bm{h}_t = &~ [\bm{h}_t^d ; \bm{h}_t^g(st) ; \bm{h}_t^g(da)] \\
% \bm{p}_j^{t+1} = &~\  \mathrm{softmax}(\mathrm{GRU}(\bm{h}_t, \bm{w}_{j-1}^{t+1})).
% \end{split}
% \end{equation*} % TILL THIS
% Given the target utterance $u_{t+1}^{'}$, we use the cross entropy loss as the loss for the generation task:
% \begin{equation}
% \label{eqn:loss3}
% \m{L}_{NLG} = - \sum_{w_j \in u^{'}_{t+1}} log(p_{j,w_j}^{t+1}).
% \end{equation}
% \textcolor{blue}{We also pass the final enriched conversation representation $\bm{h}_t$ through a linear layer to predict the negotiation success, which is denoted by the sale-to-list ratio $r = (\small{\text{sale price} - \text{buyer target price}}) / (\small{\text{listed price} - \text{buyer target price}})$ \citep{yiheng_sigdial}. 
% We split the ratios for the conversations into 5 negotiation classes of equal sizes using the training data and use those to predict the success of negotiation.} %\reminder{how to explain better?}

\subsection{model training}
\label{sec:sale_price_ratio}
%\noindent \textbf{Model Training: }
We use $\bm{h}_t^{ST}$ to predict the next set of strategies $ST_{t+1}$, a binary value vector which represents the k-hot representation of negotiation strategies for the next turn.
We compute the probability of the $j^{th}$ strategy occurring in $u_{t+1}$ as  \textcolor{black}{$\bm{p}(st_{t+1, j}| \bm{h}^{ST}_t) =  \sigma(\bm{h}_t^{ST})$.} %:
% \begin{equation*}
% %\begin{split}
% \label{eqn:one}
% %h_t^{st} = ~~ W_j[h_t^g(st)] + b_j ~~ ; ~~
% \bm{p}(st_{t+1, j}| \bm{h}^g_t(st)) =  \sigma(\bm{h}_t^{ST}),
% %end{split}
% \end{equation*}
where $\sigma$ denotes the sigmoid operator. We threshold the probability by $0.5$ to obtain the k-hot representation. We denote the weighted negative log likelihood of strategies $\bmm{L}_{ST}$ as the loss function of the task of next strategy prediction \textcolor{black}{$\bmm{L}_{ST} = - \sum_{j} \delta_j \log(\bm{p}(st_{t+1,j})) - \sum_{k} \log(1 - \bm{p}(st_{t+1,k}))$ }%:
% \begin{equation}
% \label{eqn:loss1}
% \bmm{L}_{ST} = - \sum_{j} p_j \log(st_{t+1,j}) - \sum_{k} \log(1 - st_{t+1,k}),
% \end{equation}
where the summation of $j$ are over the strategies present ($st^{'}_{t+1,j}=1$) and not present ($st^{'}_{t+1,k}=0$) in the ground truth strategies set, $ST^{'}$. Here $\delta_j$ is the positive weight associated with the particular strategy. We add this weight to the positive examples to trade off precision and recall. We put $\delta_j = \text{\small{\# of instances not having strategy j}} / \text{\small{\# of instances having strategy j}}$.

Similarly, we use $\bm{h}_t^{da}$ to predict the dialogue act for the next utterance $da_{t+1}$. Given the target dialogue act $da_{t+1}^{'}$ and the class weights $\rho_{da}$ for the dialogue acts, we denote the class-weighted cross entropy loss over the set of possible dialogue acts, %\textcolor{black}{$\bmm{L}_{DA} = - \rho_{da} \sum_{da} da_{t+1}^{'}\log(\mathrm{softmax}(\bm{h}_t^{da}))$ .}
\textcolor{black}{$\bmm{L}_{DA} = - \rho_{da} \log(\mathrm{softmax}(\bm{h}_t^{da}))$ .}
%:
% \begin{equation}
% %\begin{split}
% \label{eqn:two}
% %h_t^{da} = & ~~W[h^g_t(da)] + b\\
% %\m{L}_{DA} = &- \sum_{y'} log(\frac{\operatorname{exp}(h_t^{da}[y^{'}])}{\sum_j\operatorname{exp}(h_t^{da}[j])})
% \bmm{L}_{DA} = - w_{da} \sum_{y'} y\log(\mathrm{softmax}(\bm{h}_t^{da}))
% %\end{split}
% \end{equation}
We pass $\bm{h}_t =~ [\bm{h}_t^u ; \bm{h}_t^{ST} ; \bm{h}_t^{da}]$ through a linear layer to predict the negotiation success, which is denoted by the sale-to-list ratio $r = (\small{\text{sale price} - \text{buyer target price}}) / (\small{\text{listed price} - \text{buyer target price}})$  \citep{yiheng_sigdial}. 
We split the ratios into 5 negotiation classes of equal sizes using the training data and use those to predict the success of negotiation. %\reminder{how to explain better?}
Therefore, given the predicted probabilities for target utterance $u_{t+1}^{'}$ from \Sref{sec:decoder}, target ratio class $y_r^{'}$ and the learnable parameters $W_r$ and $b_r$, we use the cross entropy loss as the loss for the generation task ($\bmm{L}_{NLG}$) as well as the negotiation outcome prediction task ($\bmm{L}_{R}$), thus \textcolor{black}{$\bmm{L}_{NLG} = - \sum_{w_j \in u^{'}_{t+1}} \log(\bm{p}^{w_j}_{t+1})$ and $\bmm{L}_{R} = -\sum_{r \in [1,5]} y_r^{'}\log(\operatorname{softmax}(W_r\bm{h}_t + b_r))$}. The $\bmm{L}_{R}$ loss optimizes for encoding negotiation strategies to enable accurate prediction of negotiation outcome.

%%%%:
% \begin{equation}
% \label{eqn:loss3}
% \bmm{L}_{NLG} = - \sum_{w_j \in u^{'}_{t+1}} \log(\bm{p}_{j,w_j}^{t+1})~~~~~~   ~~~ \bmm{L}_{R} = -\sum_{j \in [1,5]} y_r\log(\operatorname{softmax}(W_r\bm{h}_t + b_r)).
% \end{equation}
% \begin{equation} % commented this
% \label{eqn:loss3}
% \begin{split}
% \bmm{L}_{NLG} =& - \sum_{w_j \in u^{'}_{t+1}} \log(\bm{p}_{j,w_j}^{t+1})\\
% \bmm{L}_{R} =& -\sum_{r \in [1,5]} y_r\log(\operatorname{softmax}(W_r\bm{h}_t + b_r)).
% \end{split}
% \end{equation}
%Finally, we use the cross entropy loss for this success prediction task denoted by $\bmm{L}_{R}$.
We use hyperparameters $\alpha$, $\beta$ and $\gamma$ to optimize the joint loss $\bmm{L}_{joint}$, of strategy prediction, dialogue act prediction, utterance generation and outcome prediction together, using the Adam optimizer \citep{adam_optimizer}, to get \textcolor{black}{$\bmm{L}_{joint} = \bmm{L}_{NLG} + \alpha \bmm{L}_{ST} + \beta \bmm{L}_{DA} + \gamma \bmm{L}_{R}$.}%:
% \begin{equation*}
% \bmm{L}_{joint} = \bmm{L}_{NLG} + \alpha \bmm{L}_{ST} + \beta \bmm{L}_{DA} + \gamma \bmm{L}_{R}.
% \end{equation*}

% \begin{table}[t]
%     %%%% \centering
%     \caption{Here we describe some of the dataset statistics of the CraigslistBargain dataset. We also report the max and average number of graph nodes and edges.}
%     \label{tab:appendix_dataset_stats}
%     %\scriptsize
%     \begin{center}
%     \small
%     \begin{tabular}{lc}
%     \toprule
%     Feature & Value\\ 
%     \midrule
%     Max no. of turns in any conversation & 47\\
%     Avg no. of turns & 9.2\\
%     Max no. of strategies in an utterance & 13\\
%     Avg no. of strategies per utterance & 3\\
%     Max no. of nodes in graph (total strategies) & 86\\
%     Avg no. of nodes in graph & 21\\
%     Max no. of edges in graph & 3589\\
%     Avg no. of edges in graph & 308.12\\
%      \bottomrule
%     \end{tabular}
%     \end{center}
    
% \end{table}

\section{Experimental Setup}
\label{sec:experimental_setup}

\paragraph{Dataset: }
\label{sec:dataset}
We use the CraigslistBargain dataset\footnote{\url{https://github.com/stanfordnlp/cocoa/tree/master/craigslistbargain}} \citep{negotiation_og}  to evaluate our model. The dataset was created using Amazon Mechanical Turk (AMT) in a negotiation setting where two workers were assigned the roles of buyer and seller respectively and were tasked to negotiate the price of an item on sale.% scraped from \url{craigslist.com}. 
%textcolor{red}{They were randomly assigned scenarios scraped from \url{craigslist.com} which included a product description, photos, and listing price.} 
The buyer was additionally given a target price. 
Both parties were encouraged to reach an agreement while each of the workers tried to get a better deal. 
%The raw dataset consists of 5383 train, 643 validation, and 656 test conversations. 
We remove all conversations with less than 5 turns. % to get 4828 train, 561 validation, and 567 test conversations. 
Dataset statistics are listed in \Tref{tab:appendix_dataset_stats} %along with the number of nodes and edges in the strategy graphs in Tables  \ref{tab:appendix_dataset_stats} and \ref{tab:graph_stats_table} 
in the Appendix. %\Aref{sec:appendix_logistics}. 
%The maximum number of nodes and edges in any of our strategy graphs is 86 and 3589 respectively.}%%%%, while the average numbers are 21 and 309 respectively.}
%The average number of nodes and edges in all of the strategy graphs are 21 and 309, respectively.%%%%, while the average numbers are 21 and 309 respectively.}

We extract from the dataset the coarse dialogue acts as described by \citet{negotiation_og}. This includes a list of 10 \textit{utterance dialogue acts}, e.g.,  
\textit{inform}, \textit{agree}, \textit{counter-price}.
%\textit{inform}, \textit{agree}, \textit{intro}, \textit{counter-price}, \textit{vague-price}, \textit{inquiry}, \textit{unknown}, \textit{insist}, \textit{disagree} and \textit{init-price}. 
We augment this list by 4 \textit{outcome dialogue acts}, namely, $\langle$\textit{offer}$\rangle$, $\langle$\textit{accept}$\rangle$, $\langle$\textit{reject}$\rangle$ and $\langle$\textit{quit}$\rangle$, which correspond to the actions taken by the users. 
%%%%We provide the details in Appendix \ref{sec:appendix_da}.
%%%%
Negotiation strategies are extracted from the data following \citet{yiheng_sigdial}. These include 21 fine-grained strategies grounded in prior economics/behavioral science research on negotiation %which were operationalized by strategies described in psychology and behavioral science literature 
\citep{pruitt2013negotiation,bazerman1993negotiating,bazerman2000negotiation,fisher2011getting,lax20063,thompsonpaper}, e.g, \textit{negotiate side offers}, \textit{build rapport}, \textit{show dominance}. 
%\citet{yiheng_iclr} gave a hybrid linguistic-rule and classifier based model to extract 21 actionable-negotiation strategies from an utterance, grounded in the 15 theory-based strategies. 
%\textcolor{red}{For instance, they used presence and absence of \textit{Friends} category words from LIWC \citep{liwc} to indicate the strategy \textit{build rapport}.} 
All dialogue acts and strategies are listed in Appendices~\ref{sec:appendix_da} and~\ref{sec:strategies_details_appendix}.
%In our experiments, we use the 21 negotiation strategies used by \citet{yiheng_iclr} such as \textit{first person singular}, \textit{positive sentiment}, \textit{third person singular}, \textit{hedge}, \textit{personal concern}, \textit{propose}, \textit{greet}, \textit{assertive tone}, \textit{negative sentiment}, \textit{gratitude}, \textit{first person plural}, \textit{using certainty words}, \textit{informal tone}, \textit{third person plural}, \textit{trade-in}, \textit{family} and \textit{friend}. %\rj{reviewer may raise that in ICLR paper they mentioned 15. ??}

%%%%We provide common dataset statistics including the average number of nodes and edges in the strategy graphs in Table \ref{tab:appendix_dataset_stats} in Appendix \ref{sec:appendix_logistics}. 
%%%%The maximum number of nodes and edges in any of our strategy graphs is 86 and 3589 respectively, while the average numbers are 21 and 309 respectively.

%\paragraph{Baselines: }
\label{sec:baselines}
%our main contribution, graph-based \structureencoder{} of 
\noindent \textbf{Baselines: }
\textbf{\method{}} refers to our proposed method.
To corroborate the efficacy of  \method{}, we compare it against our implementation of the present state-of-the-art model for the negotiation task: FST-enhanced hierarchical encoder-decoder model (\textbf{FeHED}) \citep{yiheng_iclr} which utilizes FSTs for encoding sequences of strategies and dialogue acts.\footnote{We replace the utterance encoder with BERT for fair comparison. This improved slightly the performance of the FeHED model compared to results published in \citet{yiheng_iclr}.} We also conduct and ablation study, and  evaluate the variants of \method{} with different ways of encoding negotiation strategies, namely, \textbf{HED}, \textbf{HED+RNN}, and \textbf{HED+Transformer}. HED completely ignores the strategy and dialogue act information, whereas HED+RNN and HED+Transformer encode them using RNN and Transformers \citep{transformer} respectively. While HED+RNN is based on the dialogue manager of \citet{negotiation_og}, HED+Transformer has not been proposed earlier for this task. 
%Since the problem of predicting strategy sequence is similar to language modeling, therefore, we use the state-of-the-art language model Transformers as an additional baseline.
For a fair comparison, we use a pre-trained BERT \citep{bert} model as the utterance encoder (\Sref{sec:model_encoder}) and a common utterance decoder (\Sref{sec:sale_price_ratio}) in all the models, and only vary the \structureencoder{}s as described above.
The strategies and dialogue acts in RNN and Transformer based encoders are fed as sequence of $k$-hot vectors.

\noindent \textbf{Evaluation Metrics: }
\label{sec:evaluation_metrics}
For evaluating the performance on the next strategy prediction and the next dialogue act prediction task, we report the F1 and ROC AUC scores for all the models. For these metrics, macro scores tell us how well the model performs on less frequent strategies/dialogue acts and the micro performance tells us how good the model performs overall while taking the label imbalance into account. Strategy prediction is a multi-label prediction problem since each utterance can have multiple strategies. 
For the downstream tasks of utterance generation, we compare the models using BLEU score \citep{bleu_paper} and BERTScore \citep{bert_score}. Finally, we also evaluate on another downstream task of predicting the outcome of negotiation, using the ratio class prediction accuracy (RC-Acc) (1 out of 5 negotiation outcome classes, as described in \Sref{sec:sale_price_ratio}). Predicting sale outcome provides better interpretability over the progression of a sale and potentially control to intervene when negotiation has a bad predicted outcome. Additionally, being able to predict the sale outcome with high accuracy shows that the model encodes the sequence of negotiation strategies well.

\section{Results}
\label{sec:results}
We evaluate (1) strategy and dialogue act prediction (intrinsic evaluation), and (2) dialogue generation and negotiation outcome prediction (downstream evaluation).
%We evaluate our model on strategy and dialogue act prediction \textcolor{blue}{which is our main focus, along with} downstream dialogue generation and quality of dialogue response generation in addition to negotiation sale price outcome, as judged by humans. 
For all metrics, we perform bootstrapped statistical tests \citep{taylor_significance,philipcohen_significance} and we bold the best results for a metric in all tables (several results are in bold if they have statistically insignificant differences). 
% \begin{itemize}[itemsep=2pt,topsep=2pt,parsep=1pt,partopsep=0pt]
% 	\item[Q1.] How much does graph-based encoder help improve the structural sequence prediction? (Sec. \ref{sec:next_strategy_prediction})	
% 	\item[Q2.] How effective is \method{} for the downstream tasks? (Sec. \ref{sec:downstream})
% 	\item[Q3.] How well does \method{} perform according to humans as compared with baselines? (Sec. \ref{sec:humaneval})
% \end{itemize}		

%\paragraph{Strategy and Dialogue Act Prediction}
\noindent \textbf{Strategy and Dialogue Act Prediction: }
%\label{sec:next_strategy_prediction}
We compare \method{}'s effectiveness in encoding the explicit sequence of strategies and dialogue acts with the baselines, using the metrics described in \Sref{sec:experimental_setup}.  
%\reminder{ add this result } We also compare the directed and undirected variants of the sequence graph. 
\Tref{tab:next_strategy_prediction_result} 
shows that \method{} performs on par with the Transformer based encoder in strategy prediction macro scores and outperforms it on other metrics. 
Moreover, both significantly outperform the FST-based based method, prior state-of-the-art. 
%We find that there is not much gain over RNNs for Dialogue Acts. 
We hypothesize that lower gains for dialogue acts are due to the limited structural dependencies between them. 
Conversely, we validate that for negotiation strategies, RNNs are significantly worse than \method{}. We also observe that higher macro scores show that \method{} and Transformers are able to capture the sequences containing the less frequent strategies/dialogue acts as well. These results supports our hypothesis of the importance to encode the structure in a more expressive model.
Moreover, \method{} also provides interpretable structures which the other baselines do not. We will discuss these findings in \Sref{sec:interpretations}.
\begin{table*}[t]
    \caption{\label{tab:next_strategy_prediction_result} Performance of the next strategy and dialogue-act prediction of various models. We report the F1 and ROC AUC scores. %Dialogue act prediction is a multi-class, one-vs-rest problem and so it won't have Micro ROC AUC.
    Significance tests were performed as described in \Sref{sec:results} and the best results (along with all statistically insignificant values) are bolded. }%FeHED is our implementation of %\citet{yiheng_iclr} 
	%where we replace the utterance encoder with BERT for a fair %comparison.}
	%\centering
	\vspace{-2mm}
	\begin{center}
	%\small
	\resizebox{\linewidth}{!}{
		\begin{tabular}{lccccccccccc}
			\toprule
			& \multicolumn{6}{c}{\textbf{Negotiation Strategies}} & \multicolumn{5}{c}{\textbf{Dialogue Acts}}\\
			\cmidrule(r){2-7} \cmidrule(r){8-12} 
			& \multicolumn{3}{c}{F1} & \multicolumn{3}{c}{ROC AUC} & \multicolumn{3}{c}{F1}  & \multicolumn{2}{c}{ROC AUC}\\
			\cmidrule(r){2-4} \cmidrule(r){5-7} \cmidrule(r){8-10} \cmidrule(r){11-12}
			\multicolumn{1}{l}{\textbf{Model}}   & Macro & Micro & Weighted & Macro & Micro & Weighted & Macro & Micro & Weighed & Macro & Weighed\\
			\midrule
			FeHED &17.6&	25.6&	36.3&	55.8&	61.7&	54.7&	20.6&	37.4&	30.6&	76.9&	79.2\\
			HED+RNN &23.2	&26.7	&42.4&	65.3&	65.3&	60.4&33.0&	\textbf{46.2}&	42.8	&83.1	&84.2	\\
			HED+Transformer 	&	\textbf{26.3}&	32.1&	\textbf{43.3}	&\textbf{68.2}&	71.8&	\textbf{61.8}&32.5&	44.6	&42.0	&\textbf{85.6}&	85.1\\
			\midrule
			\textbf{\method{}} &	\textbf{26.1}	&\textbf{34.1}&	\textbf{43.5}&	\textbf{68.1}&	\textbf{73.0}&	\textbf{61.8}&	\textbf{33.4}& \textbf{45.8}	&\textbf{43.7}	&\textbf{85.6}	&\textbf{85.4}\\
			\bottomrule
		\end{tabular}
	}
	
	\end{center}
\end{table*}

% (The FST are claimed to be interpretable but it was hard to interpret the states which just give a probability distribution over the strategies. Also, FST's can't be used to understand the importance of past strategies, on the prediction).%We show detailed Precision, Recall and F1 values for separate tactics obtained using the Transformer based and Graph based methods in Section \ref{sec:detailed_prf1} of the Appendix. 

%\begin{table}[!t]
%	\centering
%	\small
%	\resizebox{0.9\linewidth}{!}{
%		\begin{tabular}{lcc}
%			\toprule
%			\multicolumn{1}{l}{\textbf{Model}} & \multicolumn{1}{c}{\textbf{Val-F1}} & \multicolumn{1}{c}{\textbf{Test-F1}}\\
%			\midrule
			%FST & 0.20 & 0.17\\
%			FST & 0.3087 & 0.2590\\
%			RNN & 0.3287 & 0.2730\\
			%RNN & 0.1702 & 0.1386\\
%			Transformer & \textbf{0.3683} & \textbf{0.3054}\\
%			Graph-Undirected & 0.31 & 0.20 \\
%			Graph-Directed & \textbf{0.3661} & \textbf{0.3012}\\
%			\bottomrule
%		\end{tabular}
%	}
%	\caption{\label{tab:next_strategy_prediction_result} Evaluating the next-strategy-prediction performance of various models. We report the Macro F1 score. The model indicates the type of encoder used for next strategy prediction. We find that the graph based approach is comparable to the famous transformer based encoder. More details are provided in Section \ref{sec:next_strategy_prediction}.}
%\end{table}

\noindent \textbf{Automatic Evaluation on Downstream tasks: }
%\paragraph{Automatic Evaluation on Downstream tasks}
%\label{sec:downstream}
In this section, we analyze the impact of \method{} on the downstream task of Negotiation Dialogue based on the automatic evaluation metrics described in \Sref{sec:evaluation_metrics}.
In \Tref{tab:downstream}, we show that \method{} helps improve the generation of dialogue response. 
%We give the results on BLEU, BERTScore, and RC-Acc in Table \ref{tab:downstream}. 
Even though \method{} attains higher BLEU scores, we note that single-reference BLEU assumes only one possible response while dialogue systems can have multiple possible responses to the same utterance. %For example, to the utterance \textit{Hi!}, someone can respond as \textit{Hello!} or \textit{Hi!}, and both are valid. However, BLEU will give a score of zero if the specific utterance is not generated.
BERTScore alleviates this problem by scoring semantically similar responses equally high \citep{bert_score}.
We also find that both Transformer and \method{} have a comparable performance for negotiation outcome prediction, which is significantly better than the previously published baselines (FeHED and HED+RNN). 
A higher performance on this metric demonstrates that our model is able to encode the strategy sequence better and consequently predict the negotiation outcome more accurately.
%\method{} maintains comparable BLEU scores and is able to better condition the generator to generate utterances with the correct strategies.
%\noindent \textbf{Ablation Results:} %To analyze the effect of various components of \method{}, we evaluate various version of our model with cumulatively removed components. 
Additionally, ablation results in \Tref{tab:ablation} show that both strategy and dialogue act information helps \method{} in improving dialogue response. The difference in BERTScore F1 scores in Tables \ref{tab:downstream} and \ref{tab:ablation} arises due to different metrics chosen for early stopping. More details in \Aref{sec:appendix_logistics}.

Although, both HED+Transformer and \method{} are based on attention mechanisms, \method{} has the added advantage of having structural attention which helps encode the pragmatic structure of negotiation dialogues which in turn provides an interpretable interface.
%of using a graph based approach is that they provide a better interpretable interface. 
The components in our graph based encoder such as the GAT and ASAP layer provide strategy influence and cluster association information which is useful to understand and control negotiation systems. This is described in more detail in \Sref{sec:interpretations}. Though transformers have self attention, the architecture is limited and doesn't model the structure/dependence between strategies providing only limited understanding. Further, our results show that \method{} maintains or improves performance over strong models like Transformer and has much more transparent interpretability. We later show that \method{} performs significantly better than HED+Transformer in human evaluation.

\begin{figure}[t]
    \vspace{-3mm}
    \begin{minipage}{0.55\linewidth}
		%\begin{table}[!t]
    \captionof{table}{\label{tab:downstream} Downstream evaluation of negotiation dialogue generation and negotiation outcome prediction. The best results (along with all statistically insignificant values to those) are bolded.}
	%\centering
	%\begin{center}
	\small
	\resizebox{\linewidth}{!}{
		\begin{tabular}{lccccc}
			\toprule
& \multicolumn{4}{c}{\textbf{Generation}} & \multicolumn{1}{c}{\textbf{Outcome}}\\
			\cmidrule(r){2-5} 
			& & \multicolumn{3}{c}{\textsc{BERT}Score} & \textbf{Prediction}\\
			\cmidrule(r){3-5} \cmidrule(r){6-6} 
			\multicolumn{1}{l}{\textbf{Model}}  & BLEU & Precision & Recall & F1 & RC-Acc\\
			\midrule
			HED & 20.9	&21.8&	22.3&	22.1&	35.2\\
			FeHED& 23.7	&27.1&	26.8&	27.0&	42.3\\
			HED+RNN & 22.5	&22.9&	22.7	&22.8&	47.9\\
			HED+Transformer & \textbf{24.4}	&27.4&	\textbf{28.1}&	27.7&	\textbf{53.7}\\
			\midrule
			\textbf{\method{}} & \textbf{24.7}&	\textbf{27.8}&	\textbf{28.3}&	\textbf{28.1}&	\textbf{53.1}\\
			\bottomrule
		\end{tabular}
	}
	%\end{center}
%\end{table}
	\end{minipage}
	\hfill
	\begin{minipage}{0.42\linewidth}
% 		\centering
% 		\includegraphics[width=\textwidth]{./images/ablation-crop.pdf}
% 		\caption{\label{fig:ablation} Performance comparison of various ablated versions of \method{}. %Overall, strategy and dialogue act information help \method{} in improving downstream generation performance.
% 	}
		%		\caption{\label{fig:degree_plot} Plots of node classification accuracy vs. node degree. On $x$-axis we have quartiles of Node degree, i.e each bin has 25 \% of the samples in sorted order of Node degree. On $y$-axis we have the node classification accuracy of nodes whose node degree belong to that quantile.}
		
    % \begin{table}[ht]
    % \centering
    \captionof{table}{\label{tab:ablation} \method{} ablation analysis. This shows that all the different components provide complementary benefits. We also evaluate without BERT for comparison with previously published works.}
    \resizebox{\linewidth}{!}{
    \begin{tabular}{l|c}
        \toprule
        Model & BERT Score F1 \\
        \midrule
        \method{} & \textbf{27.4}\\
        w/o Strategy (ST) & 26.8 \\
        w/o ST, Dialogue Acts (DA) & 26.3 \\
        w/o ST, DA, BERT & 22.7\\
        \bottomrule
    \end{tabular}}
	\end{minipage} 
\end{figure}

\noindent \textbf{Human Evaluation: }
%\label{sec:humaneval}
Since automatic metrics only give us a partial view of the system, we complement our evaluation with detailed human evaluation. For that, we set up \method{} and the baselines on Amazon Mechanical Turk (AMT) and asked workers to role-play the buyer and negotiate with a single bot. 
After their chat is over, we ask them to fill a survey to rate the dialogue on how persuasive (\textit{My task partner was persuasive.}), coherent (\textit{My task partner’s responses were on topic and in accordance with the conversation history.}), natural (\textit{My task partner was human-like.}) and understandable (\textit{My task partner perfectly understood what I was typing.}) the bot was \footnote{We use the setup of \url{https://github.com/stanfordnlp/cocoa/}. Screenshots in \Aref{sec:screenshots}.}. 
Prior research in entailment has shown that humans tend to get better as they chat \citep{entrailment1,entrailment2} and so we restrict one user to chat with just one of the bots. 
%We have a question asking the user to select a particular option, and remove conversations where the user marks the wrong option. 
We further prune conversations which were incomplete potentially due to dropped connections.
Finally, we manually inspect the conversations extracted from AMT to extract the agreed sale price and remove conversations that were not trying to negotiate at all. % (ignored 30 out of 109 completed chats, out of which in 7 conversations the user rejected the offer).  

The results of human evaluations of the resulting 90 dialogues (about 20 per model) are presented in \Tref{tab:human_eval}. We find that baselines are
more likely to accept unfair offers and apply inappropriate strategies. Additionally, \method{} bot attained a significantly higher Sale Price Ratio, which is the outcome of negotiation, showing that effectively modeling strategy sequences leads to more effective negotiation systems. Our model also had a higher average total number of turns and words-per-turn (for just the bots) compared to all baselines, signifying engagement. It was also more persuasive and coherent while being more understandable to the user. 
%We provide some examples of AMT chats in Figure \ref{fig:example_conv}. \reminder{change this section acc to table of example conversations}% in the Appendix. 
From qualitative inspection we observe that the HED model generates utterances that are shorter and less coherent. They are natural responses like ``\textit{Yes it is}", but generic and contextually irrelevant. We hypothesize that this is due to the HED model not being optimized to encode the sequence of negotiation strategies and dialogue acts. 
 We believe that this is the reason for the high natural score for HED. From manual inspection we see that HED is not able to produce very persuasive responses.
We provide an example of a dialogue in \Aref{sec:appendix_example_conv}. We see that although HED+Transformer model performs well, \method{} achieves a better sale price outcome as it tries to repeatedly offer deals to negotiate the price.
We see that the HED is unable to understand the user responses well and tends to repeat itself.
Both the FeHED and HED baselines tend to agree with the buyer's proposal more readily whereas HED+Transformers and \method{} provide counter offers and trade-ins to persuade the user.
%We observed that \method{} was more `stubborn' than the baselines and less likely to accept early offers.  \reminder{rephrase}
%We observe that \method{} tries to repeatedly offer deals to negotiate the price. In contrast, we see that other baseline, namely transformers, readily accept deals.

\begin{figure*}[!t]
	%\centering
	\begin{center}
	\includegraphics[width=\textwidth]{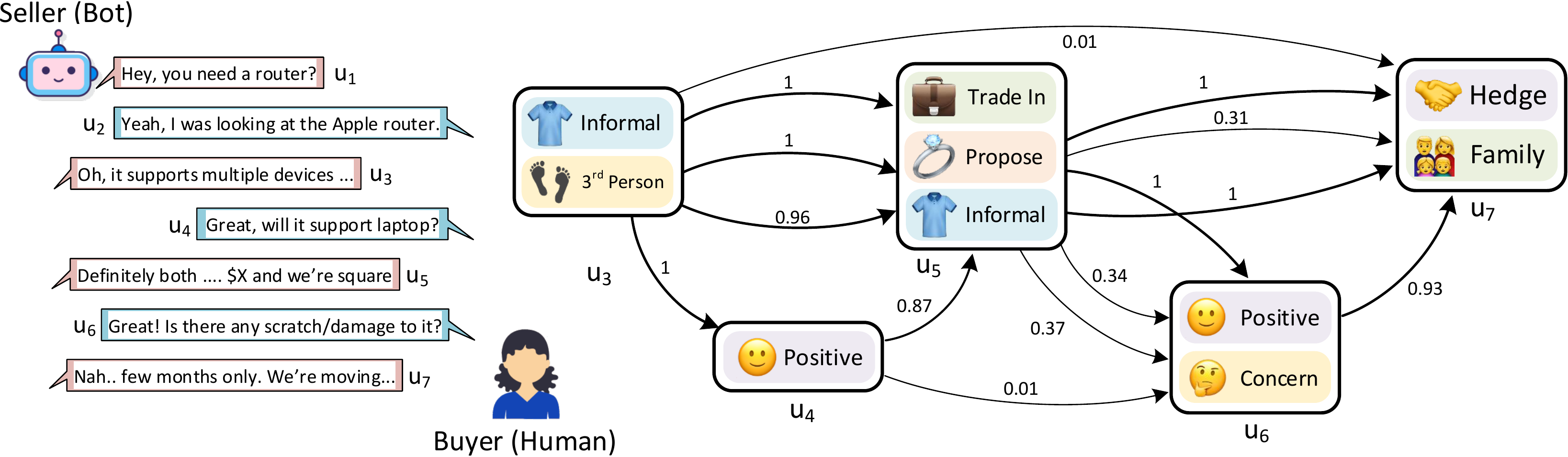}
	\end{center}
% 	\caption{\label{fig:gat_graph_figure} Visualization of the dependence attention scores as obtained from GAT where bolder edges represent higher attention. Here we present only a few edges for brevity and perform re-scaling (min-max normalization) of attention values for all incoming edges of a node. This is done because the range of raw attention values would differ based on the number of edges and this allows us to normalize any difference in scales \citep{minmax1,minmax2} and visualize the relative ranking of strategies. 
% 	%to visualize the impact of strategies in a ranked manner.
% 	For example, for \emph{trade in} at $u_5$, \emph{informal} of $u_3$ has the most influence followed by \emph{greet}.
\vspace{-2mm}
\caption{\label{fig:gat_graph_figure} Visualization of the learnt latent strategy sequences in \method{} where bolder edges represent higher influence. Here we present only a few edges for brevity and visualize min-max normalized attention values as edge weights to analyze the relative ranking of strategies.
	%to visualize the impact of strategies in a ranked manner.
	For example, for \emph{family} at $u_7$, \emph{informal} of $u_5$ has the most influence followed by \emph{propose}. 
	We present the full attention map for this example in \Fref{fig:full_attention_map} in the Appendix.%\reminder{change figure utterance to correspond better with strategies}
}
\end{figure*}
\begin{table*}[!t]
    \vspace{-2mm}
    \caption{\label{tab:human_eval} Human evaluation ratings on a scale of 1-5 for various models. We also provide the average sale price ratio (\Sref{sec:sale_price_ratio}). Negative ratio means that average sale price was lower than the buyer's target.}
	%\centering
	\vspace{-2mm}
	\begin{center}
	%\small
	\resizebox{\linewidth}{!}{
		\begin{tabular}{lcccc|ccc}
			\toprule
			\textbf{Model} & \textbf{Persuasive} & \textbf{Coherent} & \textbf{Natural} & \textbf{Understandable} & \textbf{Sale Price Ratio} & \textbf{Avg Turns} & \textbf{Avg words/turn} \\
			\midrule
			HED &2.50 & 2.50 & \textbf{4.50} & 2.50&	-2.13&11.00&4.25\\ % OLD
			%HED &3.42 & 3.85 & 4.42 & 4.42&	0.8962&11.85&4.00\\ % NEW SHIT. EARLIER RATIO WAS 0.68
			FeHED & 3.30 & 3.75 & 3.70 & \textbf{3.69} & 0.25 & 14.30 & 5.76\\ % EARLIER RATIO WAS 0.92
			HED+RNN  &2.81 & 3.27 & 3.36 & 3.27 & -3.68 & 13.90 & 3.61\\ % EARLIER RATIO WAS 0.53
			%HED+Transformer_f  	&3.8 & 4 & 4.2 & 4 & 37.13 & 11.2 & 4.47\\ % FULL DATA
			HED+Transformer  	&3.50 & 3.50 & 3.70 & 3.40 & -0.07 & 11.40 & 4.36\\ % IGNORING SPURIOUS RESPONSES - 5 ignored. 3 a chunk and 2 another % EARLIER RATIO WAS 0.89
			%HED+Transformer_1 	&3.66 & 3.66 & 3.83 & 3.58 & 36.41 & 11.25 & 4.52\\ % TAKING ONE OF THE 2 bad set of responses each. 3 ignored
			\midrule
			\textbf{\method{}}  	&\textbf{3.58} & \textbf{3.94} & 3.75 & \textbf{3.70} & \textbf{0.49} & \textbf{15.72} & \textbf{5.84}\\ % EARLIER RATIO WAS 0.95
			\bottomrule
		\end{tabular}
	}
	\end{center}
\end{table*}

\begin{table}[t]
    \caption{\label{tab:association_scores} Examples of strategies and their least / highly associated strategies based on association scores extracted using the cluster attention scores given by the ASAP layer.}
	%\centering
	\vspace{-2mm}
	\begin{center}
	%\small
	\resizebox{\linewidth}{!}{
		\begin{tabular}{l|l|l}
			\toprule
			\textbf{Negotiation Strategy} & \textbf{Least associative strategies} & \textbf{Highly associative strategies} \\
			\midrule
			concern & certainty (0.1759), trade in (0.228)& politeness please (0.7072), politeness gratitude (0.5859)\\
			%positive sentiment & trade in(0.2741), propose(0.3468)& family(0.4835), politeness greet(0.5397)\\
			hedge & trade in (0.4367), pos sentiment (0.4501) & propose (0.5427)
friend (0.6218)\\
			propose & factive count (0.3878), family (0.416)& politeness gratitude	(0.5048), trade in (0.5223)\\
			negative sentiment & trade in (0.3089), informal (0.3644) & family	(0.6363), propose (0.6495)\\
			\bottomrule
		\end{tabular}}
	\end{center}
	\vspace{-2mm}
\end{table}

\section{Interpreting Learned Strategy Graphs}
\label{sec:interpretations}
%We analyze the interpretability of \method{} at two levels, node level and graph level. 
We visualize the intermediate attention scores generated by the GATs while obtaining the strategy node representations. 
%First, we observe the attention scores given by GAT while obtaining the node (strategy) representations. 
These attention scores tell us what strategies influenced the representation of a particular strategy and can be used to observe the \textit{dependence} between  strategies \citep[cf.][]{gnn_interpret1,graph_interpret2}.
We show an example in \Fref{fig:gat_graph_figure} where for brevity, we present a subset of few turns and only the top few most relevant edges in the figure. For visualization, we re-scale the attention values for all incoming edges of a node (strategy) using min-max normalization. This is done because the range of raw attention values would differ based on the number of edges and this allows us to normalize any difference in scales and visualize the relative ranking of strategies \citep{minmax1,minmax2}. %\reminder{set up}
%We show an example in Figure \ref{fig:gat_graph_figure} %and display its full attention map in Figure \ref{fig:graph2} in Appendix Section \ref{sec:graph_examples_appendix}.
%where we notice that as soon as the propose at turn 5 happened, the strategies seem to completely change and drop dependence on previous turns.
%For brevity, we present a subset of few turns and only the top few most relevant edges in the figure.
%We display its full attention map in Figure \ref{fig:graph2} where the rows represent the turn-strategy pair which is dependent on the columns, which consist of previous turns' strategies.% in Appendix Section \ref{sec:graph_examples_appendix}.
We notice that as soon as the first \textit{propose} at $u_5$ happens, the strategies completely change and become independent of the strategies before the propose point.
%on  drop the dependence on the turns before the propose point. \reminder{verify}
From \Fref{fig:gat_graph_figure}, we see that the edge weight from $u_4$ to $u_6$ is 0.01, signifying very low influence. We noticed this trend in other examples as well, wherein, the influence of strategies coming before the first propose turn to strategies coming after that, is very low.
A similar phenomenon was also observed by \citet{yiheng_sigdial} who study the conversations by splitting into two parts based on the first propose turn.
Another interesting thing we note is that the \textit{trade-in} and \textit{propose} strategies at $u_5$ seem to be heavily influenced by \textit{informal} from $u_3$. Similarly, the \textit{informal} of $u_5$ was influenced by \textit{positive sentiment} from $u_4$. 
This indicates that the seller was influenced by previous informal interactions to \textit{propose} and \textit{trade-in} at this turn, and that sellers tend to be more informal if the conversation partner is \textit{positive}.
%We give some other examples in the Appendix in \ref{sec:appendix_example_conv}. 
In other examples, we see that at a particular utterance, different strategies depend on separate past strategies and also observe that the attention maps usually demonstrate the strategy switch as soon as the first \textit{propose} happens, which is similar to what has been observed by prior work.
These examples demonstrate that \method{} can model fine-grain strategies, learn dependence beyond just utterances and give interpretable representations, which previous baselines, including the FSTs, lack. Specifically, each state of the FST is explicitly represented by an action distribution which can only be used to see the sequence of strategies and not observe associations or dependence information which \method{} provides.

% The ASAP layer yields cluster attention scores as part of its pooling step with the aim to cluster similar graph nodes together. 
We utilize these cluster attention scores from the ASAP pooling layer to observe the \textit{association} between various strategies which can help us observe strategies with similar contextual behaviour and structural co-occurrence. 
We take the average normalized value of the cluster attention scores between two strategies to obtain the association score between them. 
%similarities across strategies and cluster together strategies which have similar contextual behaviour.
In \Tref{tab:association_scores}, we show some examples of strategies and their obtained association scores. % for a few strategies and not some interesting insights. 
%Similar to the observation by \citet{yiheng_sigdial}, negative sentiment tends to be most associated to propose.
We observe that negative sentiment tends to be most associated to propose.
We hypothesize that this is because that people who disagree more tend to get better deals. 
We observe that people do not tend to associate negative sentiment with trade-in, which is in-fact highly associated with positive sentiment,  because people might want to remain positive while offering something.
Similarly, people tend to give vague proposals by hedging, for instance, \textit{I could go lower if you can pick it up}, than when suggesting trade-in. 
Concern also seems to be least associated with certainty, and most with politeness-based strategies.
Thus, we observe that our model is able to provide meaningful insights which corroborate prior observations, justifying its ability to learn strategy associations well.

%On the graph level, we observe the cluster maps obtained from the ASAP Pooling layer. 
%These cluster attention scores are used to cluster sets of strategies as part of pooling, while learning the graph's representation. 
%They can be used to study the \textit{association} between strategies, essentially giving us information about structural co-occurrence.

\section{Related Work}
\label{sec:related_work}

\textbf{Dialogue Systems:} Goal-oriented dialogue systems have a long history in the NLP  community. 
Broadly, goal-oriented dialogue can be categorized into \emph{collaborative} and \emph{non-collaborative} systems. 
The aim of agents in a collaborative setting is to achieve a common goal, such as travel and flight reservation \citep{reservation} and information-seeking \citep{info_seeking}.
Recent years have seen a rise in non-collaborative goal-oriented dialogue systems such as persuasion \citep{persuasion4good,dutt-etal-2020-keeping,dutt2021resper}, negotiation \citep{negotiation_og,dealornodeal} and strategy games \citep{stac} due to the challenging yet interesting nature of the task.
Prior work has also focused on decision-making games such as Settlers of Catan \citep{catan} which mainly involve decision-making skills rather than communication.
\citet{dealornodeal} developed the DealOrNoDeal dataset in which agents had to reach a deal to split a set of items. 
Extensive work has been done on capturing the explicit semantic history in dialogue systems \citep{amused,vinyals2015neural,personachat}.
Recent work has shown the advantage of modeling the dialogue history in the form of belief span \citep{beliefspan} and state graphs \citep{bowden2017combiningstategraph}.
\citet{negotiation_og} proposed a bargaining scenario that can leverage semantic and strategic history.
\citet{yiheng_iclr} used unsupervisedly learned {FSTs} %Finite State Transducer
to learn dialogue structure. This approach, however, although effective in explicitly incorporating pragmatic strategies, does not leverage the expressive power of neural networks. 
Our model, in contrast, combines the interpretablity of graph-based approaches and the expressively of neural networks, improving the performance and interpretability of negotiation agents.

\textbf{Graph Neural Networks:}
%Graph Neural Networks (GNN) \citep{Bruna2013,Defferrard2016,Kipf2016} have been leveraged in several NLP applications \citep{marcheggiani-titov-2017-encoding,gcn_summ}, where they have been shown to be effective in capturing structural dependencies in various domains.
The effectiveness of GNNs \citep{Bruna2013,Defferrard2016,Kipf2016} has been corroborated in several NLP applications \citep{gnn_nlp_tutorial}, including  semantic role labeling \citep{marcheggiani-titov-2017-encoding}, machine translation \citep{bastings-etal-2017-graph},  
%multi-document summarization \citep{gcn_summ}, 
relation extraction \citep{ds_re_reside}, and knowledge graph embeddings \citep{rgcn,compgcn}. %TODO LATER
%With the advances in Graph Neural Networks (GNNs) \citep{Bruna2013,Defferrard2016,Kipf2016}, they have been extensively used in NLP applications \citep{marcheggiani-titov-2017-encoding,gcn_summ}, where they have been shown to be effective in capturing the structural dependencies in various domains. 
Hierarchical graph pooling based structure encoders have been successful in encoding graphical structures \citep{zhang2019hierarchical_graph}. 
We leverage the advances in GNNs and propose to use a graph-based explicit structure encoder to model negotiation strategies. Unlike HMM and FST based encoders, GNN-based encoders can be trained by optimizing the downstream loss and have superior expressive capabilities. 
%the expressive powers that more powerful neural-based approaches possess. 
Moreover, they provide better interpretability of the model as they can be interpreted based on observed explicit sequences \citep{graph_interpret,graph_interpret2}. 
In dialogue systems, graphs have been used to guide dialogue policy and response selection. However, they have been used to encode external knowledge \citep{dykgchat,externalKG_GAT} or speaker information \citep{dialoguegcn_speaker}, rather than compose dialogue strategies on-the-fly. Other works \citep{target,Qin_2020} focused on keyword prediction using RNN-based graphs. Our work is the first to incorporate GATs with hierarchical pooling, learning pragmatic dialogue strategies jointly with the end-to-end dialogue system. Unlike in prior work, our model leverages hybrid end-to-end and modularized architectures \citep{moss,neg_end2end} and can be plugged as explicit sequence encoder into other models.

%enforce discourse level constraints \citep{target,Qin_2020} or more commonly, 
%a graph-based approach is that it can be interpreted to observe the explicit sequences \citep{graph_interpret,graph_interpret2} which can help us study the model and understand the problem better. 
%This enables us to obtain finer levels of interpretability as compared to attention \citep{attention} based approaches, especially transformers \citep{transformer}, which are widely popular in NLP .%and can also be used to encode and interpret the various components of a sequence. %They can be used to see the importance of various utterance level signals for the next strategy prediction, whereas our graph-based approach can be used to observe intra-utterance level signals. 

\section{Conclusion}
\label{sec:conclusion}
% \textbf{Conclusion: }
We present \method{}, a novel modular negotiation dialogue system which models pragmatic negotiation strategies using Graph Attention Networks  with hierarchical pooling and learns an explicit  strategy graph jointly with the dialogue history. 
\method{} outperforms strong baselines in downstream dialogue  generation, while providing the capability to interpret and analyze the intermediate graph structures and the interactions between different strategies contextualized in the dialogue. %However, much work is still required in negotiation study to understand negotiation strategies better and improve dialogue system performance.
As future work, we would like to extend our work to discover successful (e.g.: good for the seller) and unsuccessful strategy sequences using our interpretable graph structures. 
%We believe this would play an important role in strategy recommendations for human negotiations. 
%Negotiation-coach systems can effectively use such knowledge to provide interpretable strategy recommendations for human negotiations. Our model which learns interpretable latent strategy sequences for better outcomes could directly impact such negotiation systems.
%Our proposed graph-based model can be extended with better text generation systems to generate more coherent negotiation dialogue. 
%We also plan to leverage our model for a better and interpretable negotiation-coach system. 
%This work lays the groundwork for an interpretable negotiation dialogue system that can effectively model strategy sequences and the proposed graph-based encoder can further be applied to other domains to understand structure. %We also plan to leverage our model for a better and interpretable negotiation-coach system. \reminder{end on a positive note + impact}.

\balance

\subsubsection*{Acknowledgments}
%% Ack Alissa's, Shruti's, Ritam's help, Yiheng's code, NNSA project
The authors are grateful to the anonymous reviewers for their invaluable feedback, and to Alissa Ostapenko, Shruti Rijhwani, Ritam Dutt, and members of the Tsvetshop at CMU for their helpful feedback on this work. The authors would also like to thank Yiheng Zhou for helping with negotiation strategy extraction and FeHED model.
This material is based upon work supported by the National Science Foundation under Grant No.~IIS2007960 and by the Google faculty research award. We would also like to thank Amazon for providing GPU credits.

\bibliography{iclr2021_conference}
\bibliographystyle{iclr2021_conference}

\newpage

\appendix
%\section{Appendix}
\section{Dialogue Acts}
\label{sec:appendix_da} 
Here we provide the details about the dialogue acts that we have used to annotate the utterances. 10 are taken from \citet{negotiation_og} and 4 are based on the actions taken by the users. The rule based acts are extracted using the code provided by them\footnote{\url{https://github.com/stanfordnlp/cocoa/}}. The details are in \Tref{tab:dialogue_acts}. 

\begin{table}[ht]
    
	\caption{The list of dialogue acts that we use to annotate the data.}
	%\centering
	\begin{center}
	\small
	\resizebox{\linewidth}{!}{
	    \label{tab:dialogue_acts}
		\begin{tabular}{llll}
			\hline
			\toprule
			\textbf{Meaning} & \textbf{Dialogue Act} & \textbf{Example} & \textbf{Detector} \\
			\midrule
			Greetings & intro & I would love to buy & rule\\
			Ask a question & inquiry & Sure, what's your price & rule\\
			Propose the first price & init-price & I'm on a budget so i could do \$5 & rule\\
			Proposing a counter price& counter-price & How about \$15 and I'll waive the deposit & rule\\
			Unknown & unknown & Hmm, let me think & rule\\
			Agree with the proposal & agree & That works for me & rule\\
			Disagree with a proposal & disagree & Sorry I can't agree to that & rule\\
			Answer a question & inform & This bike is brand new & rule\\
			Using comparatives with existing price & vague-price & That offer is too low & rule\\
			Insist on an offer & insist & Still can I buy it for \$ 5. I'm on a tight budget & rule\\
			Offer the price & $\langle$offer$\rangle$ & & agent action\\
			Accept the offer & $\langle$accept$\rangle$ & & agent action\\
			Reject the offer & $\langle$reject$\rangle$ & & agent action\\
			Quit the session & $\langle$quit$\rangle$ & & agent action\\
			\bottomrule
		\end{tabular}
	}
	\end{center}
\end{table}

\section{Negotiation Strategies}
\label{sec:strategies_details_appendix}

Here we provide the details about the 15 Negotiation Strategies \citep{yiheng_sigdial} and 21 Negotiation Strategies \citep{yiheng_iclr} in Tables~\ref{tab:strategies1} and \ref{tab:strategies2}. %\reminder{change this}

\begin{table*}[ht]
    \caption{The details of 15 Negotiation Strategies proposed by \citet{yiheng_sigdial}.}
	\label{tab:strategies1}
	%\centering
	\begin{center}
	\small
	\resizebox{\linewidth}{!}{
		\begin{tabular}{llll}
			\hline
			\toprule
			\textbf{High level Negotiation Rules} & \textbf{Sub Strategy} & \textbf{Example} & \textbf{Detector} \\
			\midrule
			\multirow{5}{*}{Focus on interests, not positions} & Describe Product & The car has leather seats & classifier \\
			& Rephrase product & 45k miles $\longrightarrow{}$ less than 50k miles & classifier\\
			& Embellish product & a luxury car with attractive leather seats & classifier\\
			& Address concerns & I've just taken it to maintenance & classifier\\
			& Communicate interests & I'd like to sell it asap. & classifier\\
			\hline
			\multirow{4}{*}{Invent options for mutual gain} & Propose Price & How about 9k? & classifier \\
			& Do not propose first & n/a & rule\\
			& Negotiate side offers & I can deliver it for you & rule\\
			& Hedge & I \textbf{could} come down a bit & rule\\
			\hline
			\multirow{3}{*}{Build Trust} & Communicate Politely & Greetings, gratitude, apology, please & rule \\
			& Build rapport & My kid really liked this bike, but he outgrew it & rule\\
			& Talk informally & Absolutely, ask away! & rule\\
			\hline
			\multirow{3}{*}{Insist on your position} & Show dominance & The absolute highest I can do is 640 & rule \\
			& Negative Sentiment & Sadly, I simply cannot go under 500 & rule\\
			& Certainty words & It has \textbf{always} had a screen protector & rule\\
			\bottomrule
		\end{tabular}
	}
	\end{center}
\end{table*}

\begin{table*}[ht]
	\centering
	\small
	\resizebox{0.5\linewidth}{!}{
		\begin{tabular}{lr}
			\hline
			\toprule
			\textbf{Negotiation Strategies} & \textbf{Train set frequency}\\
			\midrule
			first\_person\_singular\_count & 26,121\\
			pos\_sentiment & 24,862\\
			number\_of\_diff\_dic\_pos & 18,610\\
			third\_person\_singular & 17,000\\
			hedge\_count & 12,227\\
			number\_of\_diff\_dic\_neg & 10,402\\
			personal\_concern & 9,135\\
			propose & 8,449\\
			politeness\_greet & 6,639\\
			assertive\_count & 4,437\\
			neg\_sentiment & 3,680\\
			factive\_count & 3,429\\
			politeness\_gratitude & 3,171\\
			first\_person\_plural\_count & 2,876\\
			liwc\_certainty & 2,530\\
			liwc\_informal & 2,396\\
			third\_person\_plural & 1,721\\
			trade\_in & 883\\
			politeness\_please & 372\\
			family & 201\\
			friend & 149\\
			$<$start$>$ & 5,383\\
			\bottomrule
		\end{tabular}
	}
	\caption{The details of 21 Negotiation Strategies ($<$start$>$ added by us) used by \citet{yiheng_iclr}. These are used to operationalize the 15 strategies using a rule based system (\small{\url{https://github.com/zhouyiheng11/augmenting-non-collabrative-dialog/}}). The frequency statistics on the train set (5383 conversations) is given. A detailed description regarding the rules used by prior work to extract these are out of scope of this work, however, we intend to provide the code and extracted strategies, along with the rule based mapping to the 15 strategies upon acceptance of this work.}
	\label{tab:strategies2}
\end{table*}

\newpage

\section{Strategy-Graph Visualization}
\label{sec:appendix_graph_creation}
A visualization of a strategy sequence graph. Refer to \Sref{sec:graph_details} for more details. We also provide additional details regarding the number of nodes and edges in our strategy graphs in \Tref{tab:graph_stats_table}.

\begin{figure*}[h!]
	\centering
	\includegraphics[width=0.85\textwidth]{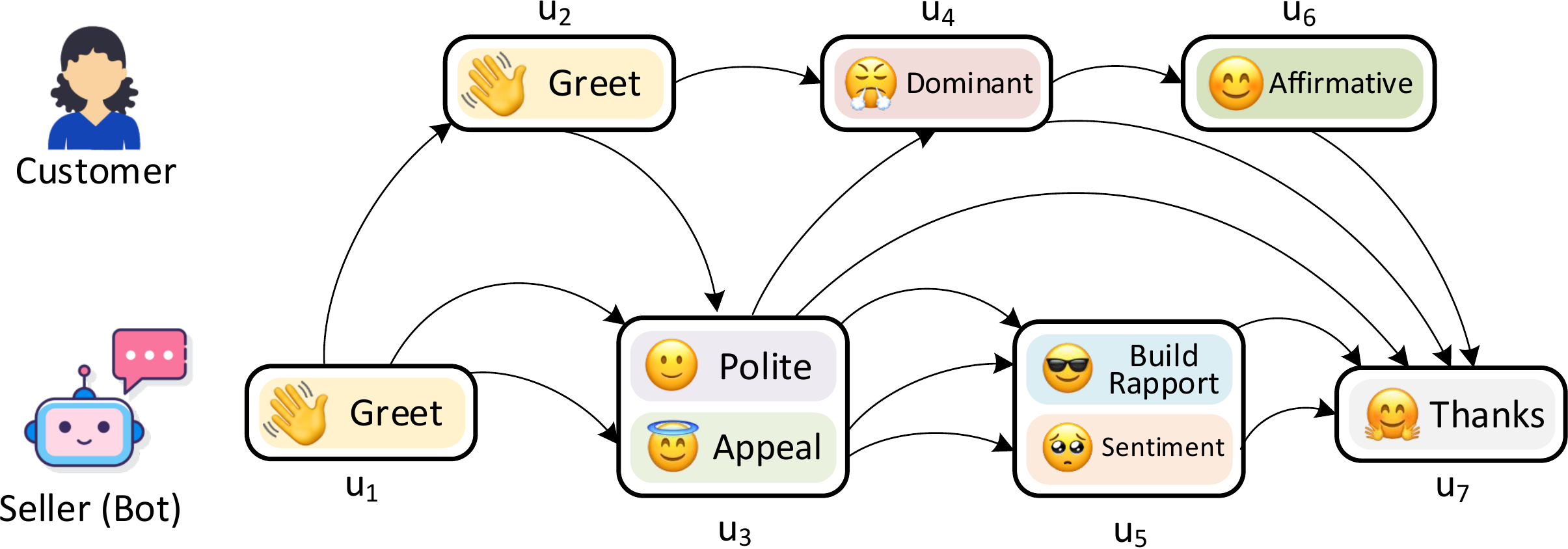}
	\caption{\label{fig:graph_creation} Visualization of a strategy sequence graph. The graph connects each strategy with all previously occurring strategies. Here we present only a few edges for brevity. For example, there would be two more additional edges from $u_4$ to the strategies of $u_5$. %\reminder{subset of edges might be misleading}
}
\end{figure*}

\begin{table}[h!]
    %%%% \centering
    \caption{We report the number of nodes and edges in our strategy-graphs. Each node corresponds to a particular utterance-strategy pair.}
    \label{tab:graph_stats_table}
    %\scriptsize
    \begin{center}
    \small
    \resizebox{0.5\textwidth}{!}{
    \begin{tabular}{lr}
    \toprule
    Feature & Value\\ 
    \midrule
    % Max no. of turns in any conversation & 47\\
    % Avg no. of turns & 9.2\\
    % \hline
    % Max no. of strategies in an utterance & 13\\
    % Avg no. of strategies per utterance & 3\\
    % \hline
    Max no. of nodes in graph (total strategies) & 86\\
    Avg no. of nodes in graph & 21\\
  %  \hline
    Max no. of edges in graph & 3589\\
    Avg no. of edges in graph & 308\\
     \bottomrule
    \end{tabular}}
    \end{center}
    
\end{table}

\newpage

\section{Hyperparameters}
\label{sec:appendix_logistics}
We present the hyper-parameters for all the experiments, their corresponding search space and their final values in \Tref{tab:hyperparameters}.  We also present additional details of our experiments below. We use most of the hyperparameters from \citet{yiheng_iclr}. Each training run took at most 3 hours on a single Nvidia GeForce GTX 1080Ti GPU and all the models were saved based on Strategy Macro F1 performance.

For experiments for Table \ref{tab:next_strategy_prediction_result} and \ref{tab:downstream} we saved the best models on best Strategy Macro F1 performance (HED being saved on outcome class prediction). This is because we wanted to prioritize and optimize our final model to capture sequence-structural information owing to our focus on interpretability.
While performing ablation studies for \Tref{tab:ablation}, not all models have structure encoders, and hence for a fair comparison we chose a metric independent of the different modules for all the models in ablations. We use the negotiation outcome class prediction (RC-Acc) scores as that optimizes the dialogue for good negotiation outcome, which indirectly helps train the model to capture the sequence of strategies. 
%Note that to perform the ablation analysis, we retrain all the models and save them based on negotiation outcome class prediction (RC-Acc) scores. 

\begin{table}[ht]
    \caption{Here we describe the search-space of all the hyper-parameters used in our experiments. }%All the experiments were run on a single 1080Ti GPU.}
    \centering
    %\scriptsize
    \small
    \resizebox{0.9\textwidth}{!}{
    \begin{tabular}{ll|c|c}
    \toprule
    Model & Hyper-parameter & Search space & Final Value\\ 
    \midrule
    All & BERT & - & bert-base-uncased no fine tuning\\
    All & BERT Dropout & - & 0.3\\
    All & Dialogue context embedding & - & 300\\
    All & Dialogue context dropout & - & 0.1\\
    All & learning-rate (lr) & 5e-3, 1e-3, 5e-4& 1e-3 \\
    All & max utterances in batch & 64,128,256 & 128 \\
    All & weighted strategy loss & True,False & True \\
    All & decay rate (l2) & - & 1e-3 \\
    All & loss alpha & 1,5 & 1 \\
    All & loss beta & - & 10 \\
    All & loss gamma & - & 10 \\
    All & projection layers for strategy & - & 64 \\
    All & projection layers for DA & - & 64 \\
    HED+RNN & hidden size & 64, 300 & 64\\
    HED+Transformer & hidden size & 64,300 & 300\\
    HED+Transformer & decoder layers & - & 6\\
    HED+Transformer & attention heads & 1,2 & 2\\
    HED+Transformer & dropout & 0.0, 0.1 & 0.0\\
    \method{} & ASAP pooling ratio & 0.2,0.5,0.8 & 0.8\\
    %\method{} & Graph model & - & GAT\\
    \method{} & hidden dim & 64,128 & 64\\
    \method{} & Graph layers & 1,2,3 & 2\\
    \method{} & Graph dropout & 0.0,0.2 & 0.0\\
     \bottomrule
    \end{tabular}}
    %\caption{Here we describe the search-space of all the hyper-parameters used in our experiments. All the experiments were run on a single 1080Ti GPU.}
    \label{tab:hyperparameters}
\end{table}

\newpage

\section{Negotiation Dataset Statistics}
In \Tref{tab:appendix_dataset_stats} we provide the CraiglistBargain dataset statistics along with data sizes after filtering conversations with less than 5 turns. The maximum and average number of turns in any conversation is 47 and 9.2 respectively. Also, the maximum and average number of strategies in an utterance is 13 and 3 respectively. 

\begin{table}[h!]
    %%%% \centering
    \caption{Dataset statistics. }
    \label{tab:appendix_dataset_stats}
    %\scriptsize
    \begin{center}
    \small
    \resizebox{0.4\textwidth}{!}{
    \begin{tabular}{lr}
    \toprule
    Data split & Size\\ 
    \midrule
    Train conversations & 5383 \\
    Valid conversations & 643 \\
    Test conversations & 656 \\
    \hline
    Filtered train conversations & 4828 \\
    Filtered valid conversations & 561 \\
    Filtered test conversations & 567 \\
    \hline
    Vocabulary size & 13339 \\
     \bottomrule
    \end{tabular}}
    \end{center}
    
\end{table}

\section{Example Conversations}
\label{sec:appendix_example_conv}
\newcommand{\humanimg}[1]{\includegraphics[width=0.025\textwidth]{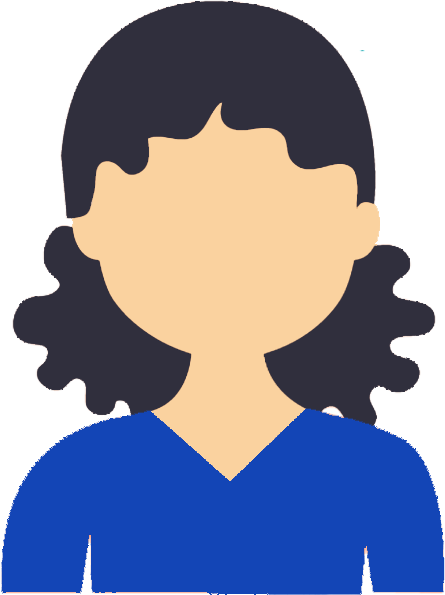}}
\newcommand{\botimg}[1]{\includegraphics[width=0.03\textwidth]{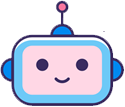}}
\begin{table*}[ht]
    \caption{Examples of the generated dialogues of various models when we keep the buyer utterances same. We see that \method{} gets the best deal for the same dialogue context and is more persistent. The FeHED and HED models accept offers more readily. We provide more examples of \method{} in \Tref{tab:more_negotiation_dialogue_example}.}
	\label{tab:negotiation_dialogue_example}
	%\centering
	\begin{center}
	\resizebox{\textwidth}{!}{
		\begin{tabular}{l||l}
			\toprule
			\multicolumn{2}{l}{\textbf{Listing Price:} 40}\\
			\multicolumn{2}{l}{\textbf{Buyer's Target Price:} 36}\\
			\multicolumn{2}{l}{\textbf{Title:} 2017 NEW Stans 24 and 26 Tubeless Tire Kit}\\
			\midrule
			\midrule
			\botimg{}: \textbf{\method{}} ~~ \humanimg{}: Human & \botimg{}: \textbf{HED+Transformer} ~~ \humanimg{}: Human\\
			\midrule
			\botimg{}: Hello & \botimg{}: Hello\\
			\humanimg{}: Hi. Is the tire kit new? & \humanimg{}: Hi. Is the tire kit new?\\
			\botimg{}: Yes it is. & \botimg{}: Yes it is.\\
			\humanimg{}: Great. My bro destroyed my old kit and I'm looking for new tires. & \humanimg{}: Great. My bro destroyed my old kit and I'm looking for new tires.\\
			\botimg{}: I am asking 40 for it. & \botimg{}: I am asking 40 for it.\\
			\humanimg{}: That's too much for a student like me. How about 35? & \humanimg{}: That's too much for a student like me. How about 35?\\
			\botimg{}: I can't go that low. I could do 38.8 and I will deliver it to you. & \botimg{}: I can't go that low. I could do 38 and throw in a warranty.\\
			\humanimg{}: I can come pick it up for 36. & \humanimg{}: I can come pick it up for 36.\\
			\botimg{}: I can do 37.2 and you have a deal. & \botimg{}: I can do that.\\
			\humanimg{}: Okay great. Thanks. $<$ offer $>$ 37.2 & \humanimg{}: Okay great. Thanks. $<$ offer $>$ 36\\
			\midrule
			\midrule
			\botimg{}: \textbf{FeHED} ~~ \humanimg{}: Human & \botimg{}: \textbf{HED} ~~ \humanimg{}: Human\\
			\midrule
			\botimg{}: Hello & \botimg{}: Hello\\
			\humanimg{}: Hi. Is the tire kit new? & \humanimg{}: Hi. Is the tire kit new?\\
			\botimg{}: Hi there! Are you interested in the bike? & \botimg{}: Yes it is.\\
			\humanimg{}: Great. My bro destroyed my old kit and I'm looking for new tires. & \humanimg{}: Great. My bro destroyed my old kit and I'm looking for new tires.\\
			\botimg{}: I am asking 40 for it. & \botimg{}: Yes it is.\\
			\humanimg{}: That's too much for a student like me. How about 35? & \humanimg{}: That's too much for a student like me. How about 35?\\
			\botimg{}: I can do that. & \botimg{}: I can do that.\\
			\humanimg{}: Okay great. Thanks. $<$ offer $>$ 35 & \humanimg{}: Okay great. Thanks. $<$ offer $>$ 35\\
			\midrule
			\midrule
			\multicolumn{2}{l}{Final Deal: Graph = 37.2, Trans = 36, FeHED = 35, HED = 35}\\
			\bottomrule
		\end{tabular}
	}
	\end{center}
\end{table*}

\begin{table*}[ht]
    \caption{More examples of the generated dialogues of \method{}.}
	\label{tab:more_negotiation_dialogue_example}
	%\centering
	\begin{center}
	\resizebox{0.52\textwidth}{!}{
		\begin{tabular}{l}
			\toprule
			\multicolumn{1}{l}{\textbf{Listing Price:} 40}\\
			\multicolumn{1}{l}{\textbf{Buyer's Target Price:} 36}\\
			\multicolumn{1}{l}{\textbf{Title:} 2017 NEW Stans 24 and 26 Tubeless Tire Kit}\\
			\midrule
			\midrule
			\botimg{} : \textbf{\method{}} ~~ \humanimg{} : Human \\
			\midrule
			\botimg{} : Hello \\
			\humanimg{} : Hi.\\
			\botimg{} : Are you interested in the kit?\\
			\humanimg{} : Yes I am interested.\\
			\botimg{} : I am looking for 40 for it.\\
			\humanimg{} : No I am willing to pay 35 for it. \\
			\botimg{} : I can't go that low \\
			\humanimg{} : What can you take?\\
			\botimg{} : I could do 38.8 and I will deliver it to you.\\
			\humanimg{} : I only have 36 please.\\
			\botimg{} : I can do 37.2 and you can pick it up yourself.\\
			\humanimg{} : Sorry i only have 36 \\
			\botimg{} : I can do that.\\
			\humanimg{} : Okay thank you.\\
			\botimg{} :  $<$ offer $>$ 37.2\\
			\midrule
			\botimg{} : Hello\\
			\humanimg{} : Hi how are you.\\
			\botimg{} : I am good.\\
			\humanimg{} : Is the kit good?\\
			\botimg{} : yes it is.\\
			\humanimg{} : are you sure?\\
			\botimg{} : yes. i am willing to negotiate the price.\\
			\humanimg{} : okay how much you willing to sell.\\
			\botimg{} : it is in great condition. i am asking 40 for it.\\
			\humanimg{} : i will give you 25.\\
			\botimg{} : i can do 38.8 and you can pick it up.\\
			\humanimg{} : 35\\
			\botimg{} : i can do 37.2 and you can pick it up.\\
			\humanimg{} : 36 last\\
			\botimg{} : ok i can do that.\\
			\humanimg{} :  $<$ offer $>$ 36\\
			\midrule
			\botimg{} : Hello\\
			\humanimg{} : Hi\\
			\botimg{} : Hi i see you are interested in my apartment\\
			\humanimg{} : i was interested in the tire kit\\
			\botimg{} : i m asking 40 for it.\\
			\humanimg{} : would you be able to do 35?\\
			\botimg{} : i m willing to negotiate a little bit. what is your budget?\\
			\humanimg{} : the highest i can go is 36 \\
			\botimg{} : i cant go that low i could do 38.8\\
			\humanimg{} : okay that works \\
			\botimg{} :  $<$ offer $>$ 38.8\\
			
			\midrule
			\midrule
		\end{tabular}
	}
	\end{center}
\end{table*}

\pagebreak
%\pagebreak 
%\pagebreak

\section{Influence Visualization}
\begin{figure*}[ht]
	\centering
	\includegraphics[width=0.85\textwidth]{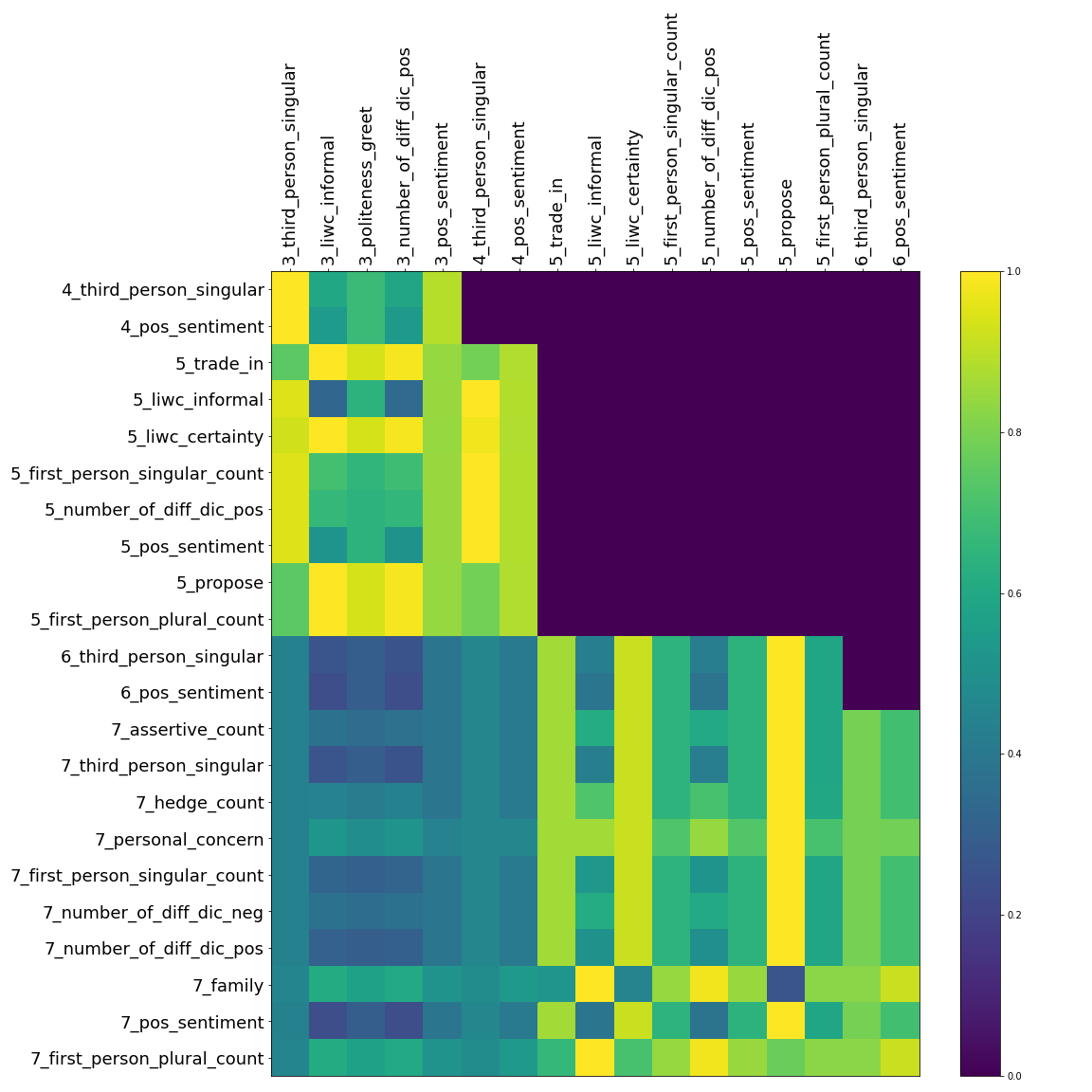}
	\caption{\label{fig:full_attention_map} Visualization of the attention map learned by \method{} for the example depicted in \Fref{fig:gat_graph_figure} in the main paper. We only show it for a few turns for brevity. Here the axis labels represent the turn and the strategy. Refer to the \Fref{fig:gat_graph_figure} in the main paper for description. %\reminder{subset of edges might be misleading}
}
\end{figure*}

Refer to \Fref{fig:full_attention_map}.

\section{Human Evaluation Interface}
\label{sec:screenshots}
\begin{figure*}[ht]
	\centering
	\includegraphics[width=0.85\textwidth]{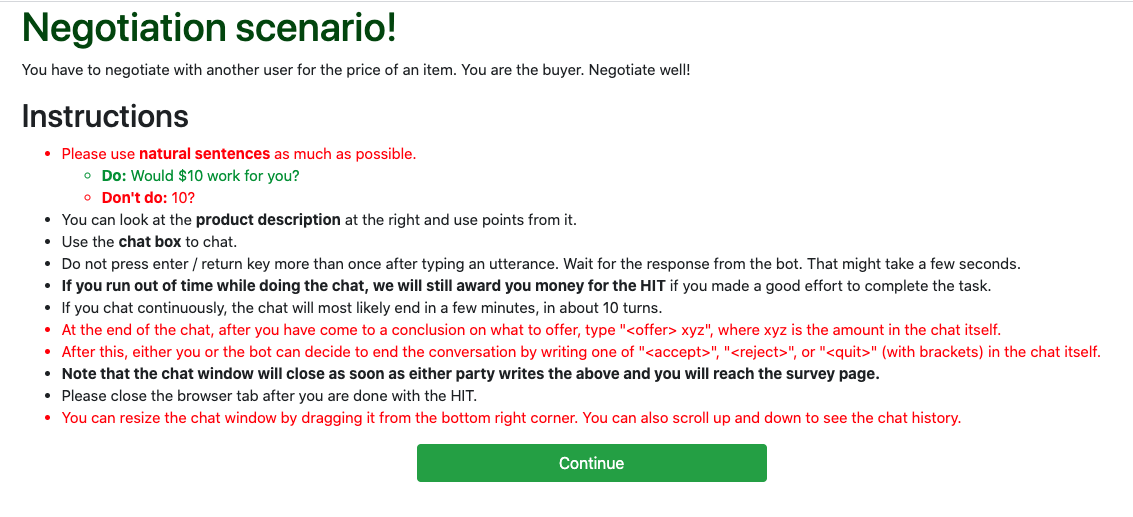}
	\caption{\label{fig:intro_screenshoot} Screenshot of the introduction for the human evaluation interface.
}
\end{figure*}
\begin{figure*}[ht]
	\centering
	\includegraphics[width=1\textwidth]{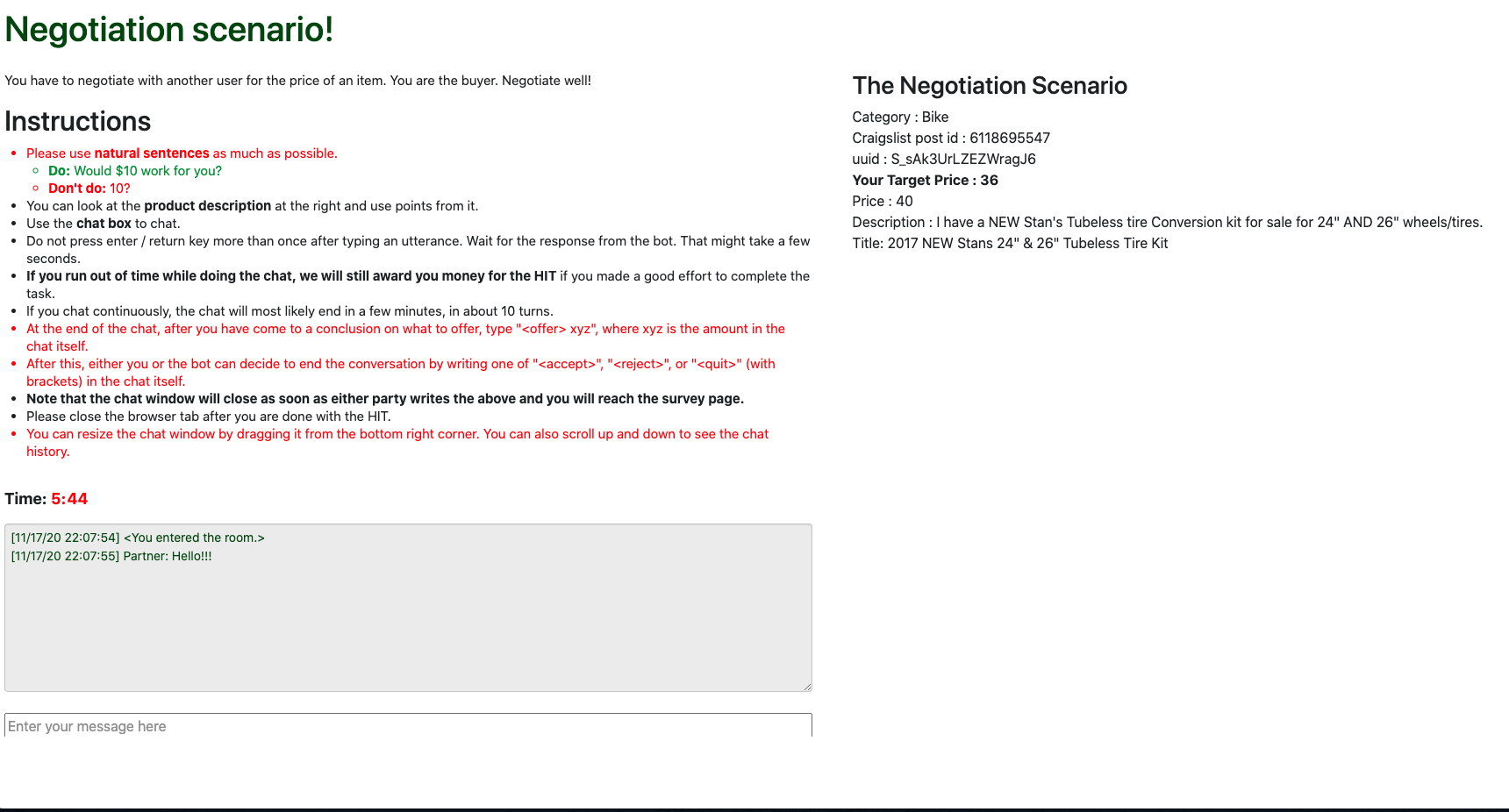}
	\caption{\label{fig:chat_window_screenshoot} Screenshot of the chat window for the human evaluation interface.
}
\end{figure*}
\begin{figure*}[ht]
	\centering
	\includegraphics[width=0.85\textwidth]{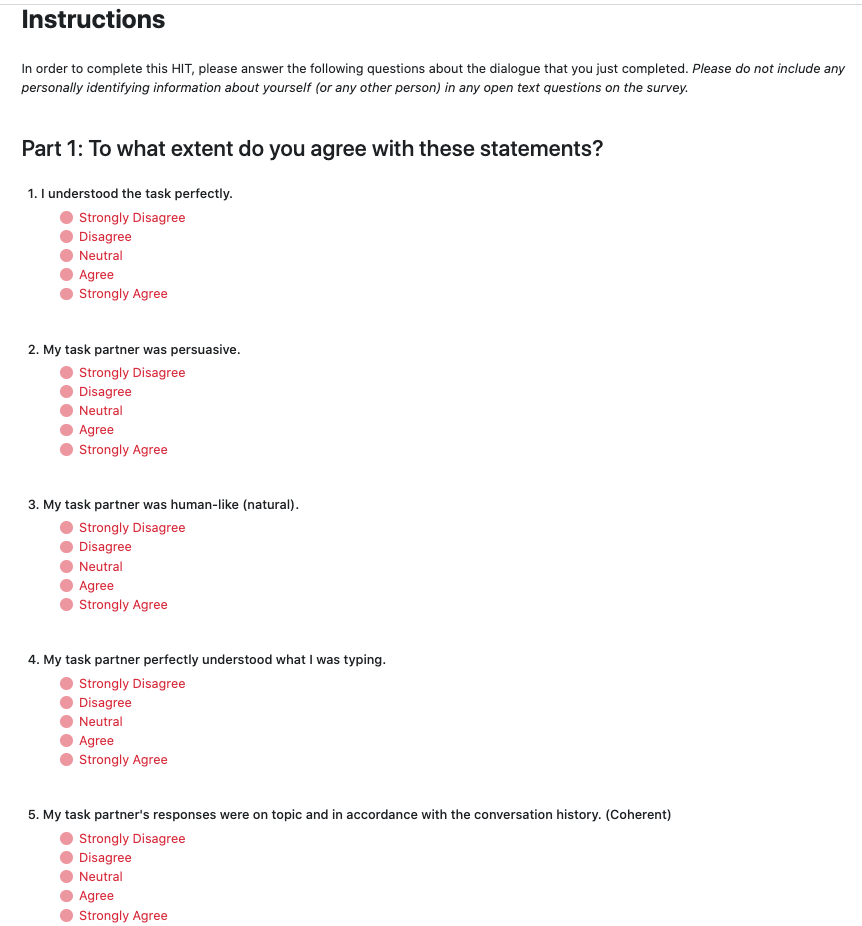}
	\caption{\label{fig:survey_screenshoot} Screenshot of the survey for the human evaluation interface.
}
\end{figure*}

%% Ack Alissa's, Shruti's, Ritam's help, Yiheng's code, NNSA project
\end{document}